\documentclass{article}

\PassOptionsToPackage{numbers, compress}{natbib}
\label{eq:opg}
\usepackage[preprint]{neurips_2023}

\usepackage[utf8]{inputenc} % allow utf-8 input
\usepackage[T1]{fontenc}    % use 8-bit T1 fonts
\usepackage{hyperref}       % hyperlinks
\usepackage{url}            % simple URL typesetting
\usepackage{booktabs}       % professional-quality tables
\usepackage{amsfonts}       % blackboard math symbols
\usepackage{nicefrac}       % compact symbols for 1/2, etc.
\usepackage{microtype}      % microtypography
\usepackage{xcolor}         % colors
\usepackage{algorithm}
\usepackage{algpseudocode}
\usepackage{amsmath,amssymb,amsthm}
\usepackage{bm}
\newtheorem{theorem}{Theorem}
\usepackage{graphicx}
\usepackage{subcaption}
\usepackage[most]{tcolorbox}
\usepackage{listings}
\usepackage{caption}
\usepackage{float}

\newtcolorbox{reasonbox}{
    enhanced,
    width=\linewidth,
    colframe=purple!80!black,
    colback=purple!5!white,
    coltitle=white,
    title filled=true,
    colbacktitle=purple!30!white,
    sharp corners=south,
    fonttitle=\bfseries,
    boxrule=0.6mm,
    left=1mm,
    right=1mm,
    top=1mm,
    bottom=1mm,
    boxsep=1mm,
    before upper=\small
}

\lstdefinestyle{pythonStyle}{
    language=Python,
    basicstyle=\ttfamily\footnotesize,
    numbers=none,
    frame=single,
    xleftmargin=0pt,
    xrightmargin=0pt,
    aboveskip=2pt,
    belowskip=2pt,
    columns=fullflexible,
    keepspaces=true
}

\title{An Expectation-Maximization Perspective on Reinforcement Learning for LLM Reasoning}

\author{%
    Tianbing Xu\\
    Aug. 2026 \\
}

\begin{document}

\maketitle

\begin{abstract}
Reinforcement learning has emerged as a powerful approach for improving the reasoning capabilities of large language models, as demonstrated by systems such as OpenAI's O1~\cite{o1} and DeepSeek-R1~\cite{r1}. However, widely used algorithms such as PPO~\cite{ppo} and GRPO~\cite{grpo} rely on likelihood-ratio correction and heuristic clipping, which increase algorithmic complexity and may introduce additional variance and optimization bias.

We present \textbf{EM Policy Gradient (EMPG)}, a simple and principled framework that formulates reinforcement learning over reasoning trajectories as an Expectation-Maximization (EM) problem~\cite{em, dempster1977em}. The E-step constructs a reward-weighted distribution over rationale trajectories, while the M-step improves the language model by maximizing their weighted log-likelihood. This formulation leads to a practical off-policy policy-gradient algorithm that directly optimizes the EM surrogate objective without trajectory-level importance weighting or likelihood-ratio clipping. Moreover, under exact E- and improving M-steps, EMPG guarantees monotonic improvement of the expected-reward objective.

Experiments on GSM8K and MATH Hard with multiple Qwen2.5 base models show that EMPG achieves performance comparable to or better than GRPO while using a substantially simpler optimization procedure. EMPG also produces more concise reasoning trajectories and exhibits structured reasoning behaviors, including subproblem decomposition, self-verification, reflection, and backtracking. These results suggest that the EM perspective provides an effective theoretical and algorithmic foundation for reinforcement learning of reasoning language models.
\end{abstract}

\section{Introduction}

Recent advances in large language models (LLMs) have demonstrated that
reinforcement learning (RL~\cite{rl}) can substantially improve reasoning
capabilities beyond supervised fine-tuning. Foundation models such as
OpenAI's O1~\cite{o1} and DeepSeek-R1~\cite{r1} exhibit remarkable
performance on mathematical reasoning, coding, and complex planning by
learning to generate intermediate reasoning processes, commonly known as
\emph{chains of thought} (CoT)~\cite{cotw}. These successes have
established reinforcement learning as a key ingredient in the emerging
paradigm of \emph{learning to reason}.

Despite their empirical effectiveness, existing reinforcement learning
algorithms for reasoning language models remain largely adaptations of
general-purpose policy optimization methods. Algorithms such as
PPO~\cite{ppo} and GRPO~\cite{grpo} optimize an on-policy objective by
correcting distribution mismatch through trajectory-level importance
weighting and stabilizing updates using heuristic clipping. While these
techniques have proven successful in practice, they increase algorithmic
complexity and may introduce additional variance and optimization bias,
particularly for long reasoning trajectories. These observations raise a
natural question:

\emph{Can reinforcement learning for reasoning be formulated from a
different optimization perspective that admits a simpler learning
algorithm?}

In this paper, we answer this question by viewing reinforcement learning
over reasoning trajectories through the lens of
Expectation-Maximization (EM)~\cite{em, dempster1977em}. We formulate the objective of
learning to reason as maximizing the expected reward over latent
reasoning trajectories and show that this objective admits a natural EM
interpretation. In the E-step, the algorithm constructs a
reward-weighted distribution over reasoning trajectories, while in the
M-step it improves the language model by maximizing their weighted
log-likelihood. This formulation separates trajectory generation from
policy optimization and provides a principled optimization framework for
reasoning language models.

Building upon this formulation, we propose
\textbf{EM Policy Gradient (EMPG)}, a practical reinforcement learning
algorithm that approximates the exact EM optimization. EMPG implements
the E-step through reward-guided trajectory sampling and the M-step
through an off-policy policy-gradient update. Unlike PPO and
GRPO, EMPG directly optimizes the EM surrogate objective without
trajectory-level importance weighting or likelihood-ratio clipping,
resulting in a substantially simpler optimization procedure.
Furthermore, we show that the exact EM formulation enjoys a monotonic
improvement guarantee: under exact E- and improving M-steps, each EM
iteration provably increases the expected-reward objective, providing a
theoretical foundation for the proposed algorithm.

We evaluate EMPG on the GSM8K and MATH Hard benchmarks using multiple
Qwen2.5 base models. Across all model scales, EMPG achieves performance
comparable to or better than the state-of-the-art GRPO algorithm while
using a considerably simpler optimization procedure. Beyond benchmark
accuracy, EMPG produces more concise reasoning trajectories and exhibits
rich reasoning behaviors, including subproblem decomposition,
self-verification, reflection, and backtracking.

Our contributions are summarized as follows:

\begin{itemize}
\item We formulate reinforcement learning for reasoning language models
as an Expectation-Maximization optimization problem, providing a new
perspective on learning over latent reasoning trajectories.

\item We derive EM Policy Gradient, a practical off-policy
policy-gradient algorithm that naturally follows from the EM
formulation, eliminating the need for trajectory-level importance
weighting and heuristic clipping.

\item We establish a monotonic improvement guarantee for the exact EM
optimization procedure and empirically demonstrate that EMPG achieves
competitive reasoning performance while substantially simplifying
optimization.
\end{itemize}

%%%%%%%%%%%%%%%%%%%%%%%%%%%%%%%%%%%%%%%%%%%%%%%%%%%%%%%%%%%%%%
\section{EM Formulation for Learning to Reason}
\label{sec:em}

We formulate reinforcement learning over reasoning trajectories as an
Expectation-Maximization (EM) optimization problem. We first introduce
the latent trajectory model and the expected-reward objective, derive a
variational lower bound, and then characterize the corresponding E- and
M-steps. Finally, we establish a monotonic-improvement guarantee for the
resulting optimization procedure.

%%%%%%%%%%%%%%%%%%%%%%%%%%%%%%%%%%%%%%%%%%%%%%%%%%%%%%%%%%%%%%
\subsection{Problem Formulation}

Let $x\sim Q(x)$ denote an input query. We model the reasoning process
as a latent rationale sequence

\[
\mathbf z
=
(z_1,\ldots,z_T),
\]

generated autoregressively by a reasoning policy

\[
\pi_\theta(\mathbf z\mid x)
=
\prod_{t=1}^{T}
\pi_\theta
(
z_t
\mid
z_{<t},x
).
\]

The trajectory length $T$ is variable, allowing the model to generate
reasoning traces of different lengths depending on the complexity of the
problem.

Conditioned on the latent rationale $\mathbf z$, the model produces the
final response $y$. We define the complete reasoning trajectory as

\[
\tau
=
(x,\mathbf z,y),
\]

whose distribution is

\begin{align}
\label{eq:trajectory}
P_\theta(\tau)
&=
Q(x)\,
P_\theta(\mathbf z,y\mid x)
=
Q(x)\,
\pi_\theta(\mathbf z\mid x)
P(y\mid x,\mathbf z).
\end{align}

In practice, both the latent rationale and the final response are
generated autoregressively by the same large language model.

Each trajectory is assigned a reward

\[
R(\tau)\ge0,
\]

which measures the quality of the generated reasoning trajectory and the
final response. For reasoning tasks with verifiable answers, such as
mathematics and code generation, the reward may be computed
automatically. More generally, it may be provided by a learned reward
model.

The objective of learning to reason is to maximize the expected reward

\begin{align}
\label{eq:l2r}
J(\theta)
&:=
\mathbb E_{\tau\sim P_\theta}
[
R(\tau)
]
%=
%\sum_\tau
%P_\theta(\tau)R(\tau).
\end{align}

%%%%%%%%%%%%%%%%%%%%%%%%%%%%%%%%%%%%%%%%%%%%%%%%%%%%%%%%%%%%%%
\subsection{Variational Lower Bound}

Directly maximizing Eq.~\eqref{eq:l2r} is difficult because the reward is
defined over complete reasoning trajectories. We therefore introduce a
variational trajectory distribution $q(\tau)$ and derive a lower bound
on the log objective.

For any distribution $q(\tau)$,

\begin{align}
\log J(\theta)
&=
\log
\sum_\tau
q(\tau)
\frac{
P_\theta(\tau)
R(\tau)
}{
q(\tau)
}
\nonumber\\
&\ge
\underbrace{
\mathbb E_{\tau\sim q}
\left[
\log P_\theta(\tau)
+
\log R(\tau)
-
\log q(\tau)
\right]
}_{\mathcal L(\theta,q)},
\label{eq:lb}
\end{align}

where the inequality follows from Jensen's inequality.

The lower bound admits the decomposition

\begin{align}
\label{eq:decomposition}
\log J(\theta)
=
\mathcal L(\theta,q)
+
D_{\mathrm{KL}}
(
q
\|
q_\theta^\star
),
\end{align}

where

\begin{align*}
\label{eq:posterior}
q_\theta^\star(\tau)
=
\frac{
P_\theta(\tau)
R(\tau)
}{
J(\theta)
}
\end{align*}

is the reward-weighted posterior trajectory distribution.

Equation~\eqref{eq:decomposition} shows that the lower bound is tight if
and only if

\[
q
=
q_\theta^\star.
\]

%%%%%%%%%%%%%%%%%%%%%%%%%%%%%%%%%%%%%%%%%%%%%%%%%%%%%%%%%%%%%%
\subsection{Expectation-Maximization Optimization}

The EM algorithm alternates between optimizing the variational
distribution and the trajectory model.

\paragraph{E-step.}

Given the current parameters $\theta_t$, the E-step computes

\begin{align*}
q_t
&=
\arg\max_q
\mathcal L(\theta_t,q)
=
q_{\theta_t}^\star,
\end{align*}

which is given by

\begin{align}
\label{eq:estep}
q_t(\tau)
=
\frac{
P_{\theta_t}(\tau)
R(\tau)
}{
J(\theta_t)
}.
\end{align}

Thus, the E-step reweights trajectories according to their rewards,
placing greater probability mass on higher-quality reasoning
trajectories.

\paragraph{M-step.}

Holding $q_t$ fixed, the M-step updates the model parameters by solving

\begin{align*}
\theta_{t+1}
=
\arg\max_\theta
\mathcal L(\theta,q_t).
\end{align*}

Since both $q_t$ and $R(\tau)$ are fixed during the M-step, maximizing
the variational lower bound is equivalent to maximizing the
reward-weighted trajectory log-likelihood

\begin{align}
\label{eq:mstep_objective}
\theta_{t+1}
=
\arg\max_\theta
\mathbb E_{\tau\sim q_t}
[
\log P_\theta(\tau)
].
\end{align}

Equations~\eqref{eq:estep} and
\eqref{eq:mstep_objective} define the exact population-level EM
optimization. In Section~\ref{sec:empg}, we develop a practical
approximation that replaces the intractable posterior distribution by
sampled reasoning trajectories while preserving the same optimization
principle.

%%%%%%%%%%%%%%%%%%%%%%%%%%%%%%%%%%%%%%%%%%%%%%%%%%%%%%%%%%%%%%
\subsection{Monotonic Improvement}

The EM formulation naturally yields a monotonic-improvement guarantee.

\begin{theorem}[Monotonic Improvement of EM Policy Gradient]
\label{thm:monotonic_improvement}

Let

\[
J(\theta)
=
\mathbb E_{\tau\sim P_\theta}
[
R(\tau)
],
\]

where $R(\tau)>0$.

Suppose that the E-step computes the exact posterior

\[
q_t
=
q_{\theta_t}^\star,
\]

and the subsequent M-step satisfies

\[
\mathcal L(\theta_{t+1},q_t)
\ge
\mathcal L(\theta_t,q_t).
\]

Then

\[
J(\theta_{t+1})
\ge
J(\theta_t).
\]

\end{theorem}

Theorem~\ref{thm:monotonic_improvement} establishes that exact EM Policy
Gradient inherits the classical monotonic-improvement property of the EM
algorithm. In practice, the E- and M-steps are approximated using finite
trajectory samples and stochastic gradient optimization. The theorem
therefore characterizes the idealized optimization procedure and
provides the theoretical foundation for the practical algorithm
developed in the next section. A complete proof is given in
Appendix~\ref{sec:supplementary_monotonic_proof}.

%%%%%%%%%%%%%%%%%%%%%%%%%%%%%%%%%%%%%%%%%%%%%%%%%%%%%%%%%%%%%%
\section{EM Policy Gradient}
\label{sec:empg}

Section~\ref{sec:em} formulates learning to reason as an exact
Expectation-Maximization optimization problem. In practice, however,
the reward-weighted posterior trajectory distribution in
Eq.~\eqref{eq:estep} cannot be sampled directly, and the exact M-step
cannot be solved analytically. This section develops a practical
approximation, called \textbf{EM Policy Gradient (EMPG)}, that
approximates the E-step by reward-guided trajectory sampling and the
M-step by a one-step off-policy policy-gradient update.

%%%%%%%%%%%%%%%%%%%%%%%%%%%%%%%%%%%%%%%%%%%%%%%%%%%%%%%%%%%%%%
\subsection{Practical E-Step}
\label{sec:practical_estep}

The exact E-step requires sampling trajectories from the posterior

\[
q_t(\tau)
\propto
P_{\theta_t}(\tau)R(\tau),
\]

which is generally intractable because the normalization constant is
unknown.

Instead, we approximate the posterior using trajectories generated by
the current policy. Given a mini-batch of input queries
\(x\sim Q(x)\), we sample multiple rationale-response trajectories

\[
(\mathbf z,y)
\sim
P_{\theta_t}(\mathbf z,y\mid x)
\]

using stochastic decoding with a relatively high sampling temperature.
Each trajectory is evaluated by the reward function \(R(\tau)\), and
the resulting tuples

\[
(x,\mathbf z,y,R(\tau))
\]

are stored in a replay buffer $\mathcal D_t$.

This procedure can be viewed as a Monte Carlo approximation of the
reward-weighted posterior, where trajectories with larger rewards are
assigned greater influence during the subsequent optimization.

%Unlike conventional Best-of-\(N\) inference, which retains only the
%highest-scoring sample, EMPG preserves all sampled trajectories together
%with their rewards. This provides substantially richer supervision,
%allowing the policy to learn from both successful and unsuccessful
%reasoning attempts.

%%%%%%%%%%%%%%%%%%%%%%%%%%%%%%%%%%%%%%%%%%%%%%%%%%%%%%%%%%%%%%
\subsection{Practical M-Step}
\label{sec:practical_mstep}

Given the replay buffer \(\mathcal D_t\), the exact M-step objective

\[
\mathbb E_{\tau\sim q_t}
[
\log P_\theta(\tau)
]
\]

is approximated by reward-weighted empirical risk minimization,

\[
L_t(\theta)
=
\frac1{|\mathcal D_t|}
\sum_{\tau\in\mathcal D_t}
R(\tau)
\log P_\theta(\tau).
\]

The corresponding gradient estimator is

\begin{align}
\label{eq:one_pg}
\nabla_\theta L_t(\theta)
=
\mathbb E_{\tau\sim\mathcal D_t}
\left[
\nabla_\theta
\log P_\theta(\tau)
R(\tau)
\right].
\end{align}

Since all trajectories are generated by the policy from the immediately
preceding EM iteration, we refer to Eq.~\eqref{eq:one_pg} as a
\emph{one-step off-policy policy gradient}.

Because the replay buffer contains only recently generated
trajectories, the distribution mismatch between the behavior policy and
the updated policy remains small in practice, while allowing efficient
trajectory reuse during optimization.

%%%%%%%%%%%%%%%%%%%%%%%%%%%%%%%%%%%%%%%%%%%%%%%%%%%%%%%%%%%%%%
\subsection{Relation to On-Policy Policy Gradient}

The original reinforcement-learning objective can be rewritten using
trajectories sampled from the previous policy,

\[
J(\theta)
=
\mathbb E_{\tau\sim P_{\theta_{\rm old}}}
\left[
\frac{
P_\theta(\tau)
}{
P_{\theta_{\rm old}}(\tau)
}
R(\tau)
\right].
\]

Differentiating yields

\begin{align}
\label{eq:onpolicy}
\nabla_\theta J(\theta)
=
\mathbb E_{\tau\sim P_{\theta_{\rm old}}}
\left[
w(\tau)
\nabla_\theta
\log P_\theta(\tau)
R(\tau)
\right],
\end{align}

where

\[
w(\tau)
=
\frac{
P_\theta(\tau)
}{
P_{\theta_{\rm old}}(\tau)
}
\]

is the trajectory-level importance weight.

The only difference between Eq.~\eqref{eq:onpolicy} and the proposed
EMPG gradient in Eq.~\eqref{eq:one_pg} is the absence of the importance
weight.

PPO and GRPO employ importance weighting together with likelihood-ratio
clipping to compensate for the mismatch between the behavior policy and
the updated policy. However, trajectory-level importance weights may
have extremely high variance for long reasoning trajectories, while
clipping introduces additional bias and restricts policy updates.

In contrast, EMPG directly optimizes the reward-weighted M-step
objective derived from the EM formulation. Consequently, it avoids both
trajectory-level importance weighting and heuristic clipping, yielding a
simpler and lower-variance optimization procedure.

%%%%%%%%%%%%%%%%%%%%%%%%%%%%%%%%%%%%%%%%%%%%%%%%%%%%%%%%%%%%%%
%%%%%%%%%%%%%%%%%%%%%%%%%%%%%%%%%%%%%%%%%%%%%%%%%%%%%%%%%%%%%%
\begin{algorithm}[t]
\caption{EM Policy Gradient}
\label{alg:empg}
\small
\begin{algorithmic}[1]

\Require Initial parameters $\theta_0$, query distribution $Q$,
reward function $R$
\Require Iterations $T$, samples per query $N$,
M-step updates $K$, learning rate $\eta$

\For{$t=0,\ldots,T-1$}

    \State $\theta^{\mathrm{old}} \gets \theta_t$
    \State $\mathcal D_t \gets \emptyset$

    \Statex \textbf{E-step: Collect reward-annotated trajectories}

    \For{each query $x\sim Q$}
        \For{$i=1,\ldots,N$}
            \State $(\mathbf z^i,y^i)
            \sim
            P_{\theta^{\mathrm{old}}}(\mathbf z,y\mid x)$
            \State $\tau^i \gets (x,\mathbf z^i,y^i)$
            \State $\mathcal D_t
            \gets
            \mathcal D_t
            \cup
            \{(\tau^i,R(\tau^i))\}$
        \EndFor
    \EndFor

    \Statex \textbf{M-step: Reward-weighted policy optimization}

    \State $\theta \gets \theta^{\mathrm{old}}$

    \For{$k=1,\ldots,K$}
        \State Sample a mini-batch $M\subseteq\mathcal D_t$
        \State Compute $\widetilde R(\tau)$ and $b_M$ for $\tau\in M$
        \State $\displaystyle
        \widehat g
        \gets
        \frac{1}{|M|}
        \sum_{\tau\in M}
        \bigl(\widetilde R(\tau)-b_M\bigr)
        \nabla_\theta\log P_\theta(\tau)$
        \State $\theta\gets\theta+\eta\widehat g$
    \EndFor

    \State $\theta_{t+1}\gets\theta$

\EndFor

\State \Return $\theta_T$

\end{algorithmic}
\end{algorithm}

\subsection{EM Policy Gradient Algorithm}
\label{sec:empg_algorithm}

Algorithm~\ref{alg:empg} summarizes the complete EM Policy Gradient
procedure. At iteration \(t\), the practical E-step freezes the current
policy \(P_{\theta_t}\), samples multiple rationale-response trajectories
for each query, evaluates their rewards, and stores the resulting
reward-annotated trajectories in a buffer \(\mathcal D_t\). The practical
M-step then performs reward-weighted fine-tuning on this fixed buffer.

\paragraph{Reward transformation.}
For reasoning tasks with verifiable answers, the raw reward is often
binary and sparse. To obtain a smoother optimization signal, we
standardize rewards within each mini-batch \(M\) and apply a positive,
monotonically increasing transformation:
\begin{align*}
\label{eq:reward_smoothing}
\widetilde R(\tau)
&=
\delta
\left(
\frac{
R(\tau)-\mu_M
}{
\sigma_M+\epsilon
}
\right),
\\
\mu_M
&=
\frac{1}{|M|}
\sum_{\kappa\in M}
R(\kappa),
\nonumber\\
\sigma_M
&=
\sqrt{
\frac{1}{|M|}
\sum_{\kappa\in M}
\left(
R(\kappa)-\mu_M
\right)^2
},
\nonumber
\end{align*}
where \(\epsilon>0\) is a numerical-stability constant and
\[
\delta(u)
=
\frac{1}{1+\exp(-u)}
\]
is the sigmoid function. This transformation maps standardized rewards
to \((0,1)\) while preserving their ordering.

\paragraph{Variance reduction.}
Following REINFORCE~\cite{reinforce}, we subtract a baseline \(b_M\)
from the transformed reward. We use the mini-batch mean
\begin{align*}
b_M
=
\frac{1}{|M|}
\sum_{\tau\in M}
\widetilde R(\tau).
\end{align*}
The practical M-step objective is
\begin{align*}
\label{eq:practical_mstep_objective}
\widehat{\mathcal M}_t(\theta;M)
=
\frac{1}{|M|}
\sum_{\tau\in M}
\left(
\widetilde R(\tau)-b_M
\right)
\log P_\theta(\tau),
\end{align*}
with gradient estimator
\begin{align*}
\widehat g_t(\theta;M)
=
\frac{1}{|M|}
\sum_{\tau\in M}
\left(
\widetilde R(\tau)-b_M
\right)
\nabla_\theta
\log P_\theta(\tau).
\end{align*}

Since the query distribution \(Q(x)\) is independent of \(\theta\),
\[
\nabla_\theta\log P_\theta(\tau)
=
\nabla_\theta
\log P_\theta(\mathbf z,y\mid x).
\]
Thus, Eq.~\eqref{eq:opg} can be implemented by applying the
reward-weighted log-likelihood loss to the generated rationale and
response tokens.

The E-step uses stochastic decoding to explore multiple reasoning paths
for each query. Unlike conventional Best-of-\(N\) inference, EMPG does
not retain only the highest-reward trajectory; instead, it preserves all
sampled trajectories and uses their rewards to determine their influence
during the M-step. This allows the model to learn from both successful
and unsuccessful reasoning attempts.

The M-step may perform multiple gradient updates on the fixed buffer
\(\mathcal D_t\). During these updates, the behavior policy remains
\(P_{\theta^{\mathrm{old}}}\), while the optimized policy
\(P_\theta\) changes. The resulting estimator is therefore off-policy,
but only with respect to trajectories generated at the immediately
preceding EM iteration. After the M-step, the updated model becomes the
behavior policy for the next E-step.

Unlike PPO and GRPO, EMPG does not require likelihood-ratio clipping,
trajectory-level importance weighting, a separate value model, or
generalized advantage estimation. Its optimization consists only of
trajectory generation, reward evaluation, and reward-weighted
language-model fine-tuning.

\begin{figure}[htbp]
    \centering
    \begin{subfigure}[b]{0.48\textwidth}
        \includegraphics[width=\textwidth]{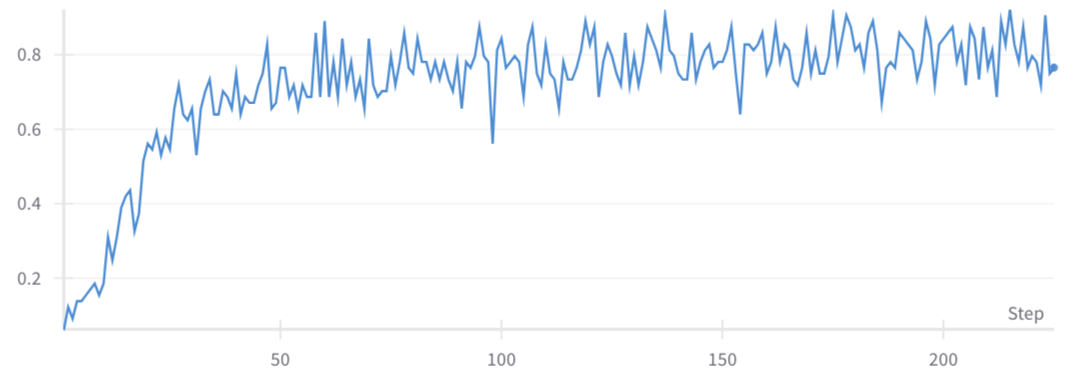}
        \caption{GSM8K}
        \label{fig:rw_gsm_1_5}
    \end{subfigure}
    \hfill
    \begin{subfigure}[b]{0.48\textwidth}
        \includegraphics[width=\textwidth]{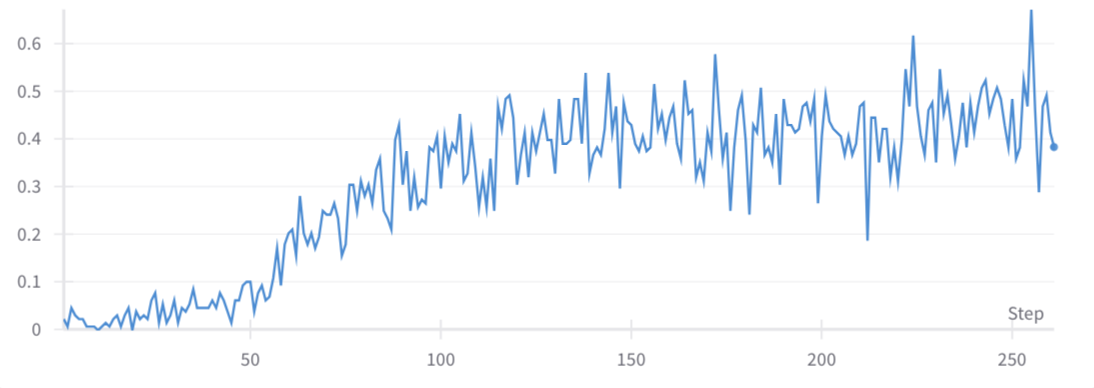}
        \caption{MATH HARD}
        \label{fig:rw_math_1_5}
    \end{subfigure}
    \caption{EMPG Learning Curves (Base: QWen2.5-1.5B)}
    \label{fig:rw_1_5}
\end{figure}

\begin{figure}[htbp]
    \centering
    \begin{subfigure}[b]{0.48\textwidth}
        \includegraphics[width=\textwidth]{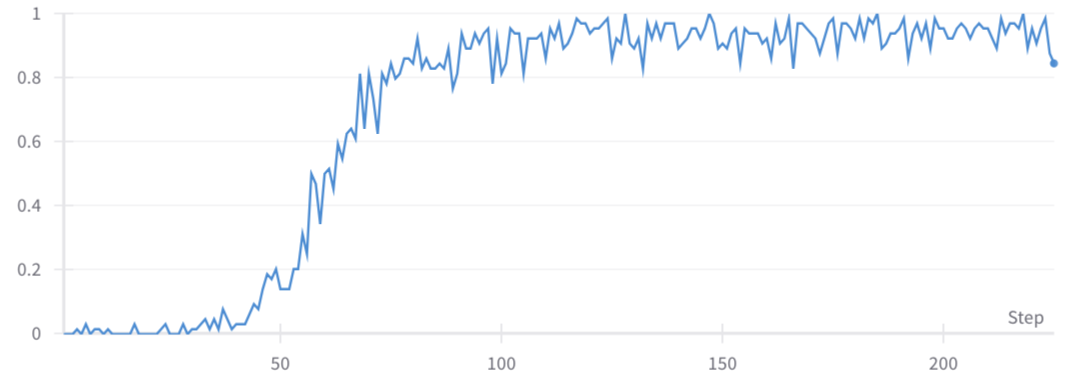}
        \caption{GSM8K}
        \label{fig:rw_gsm_7bm}
    \end{subfigure}
    \hfill
    \begin{subfigure}[b]{0.48\textwidth}
        \includegraphics[width=\textwidth]{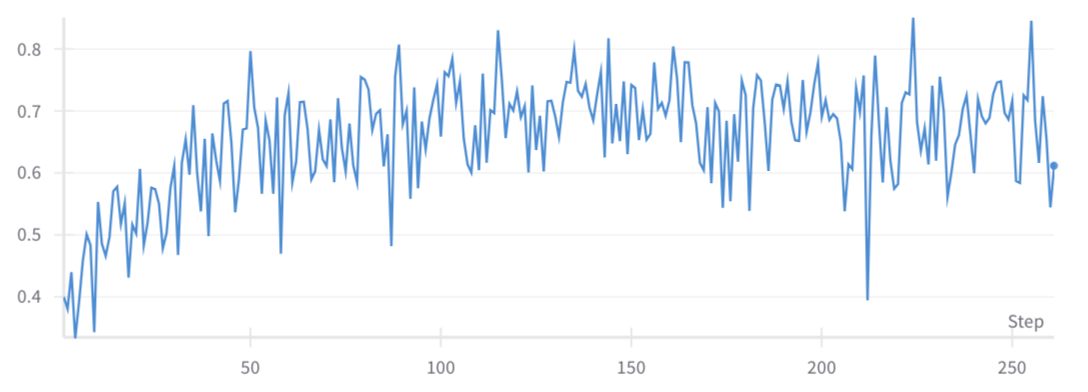}
        \caption{MATH HARD}
        \label{fig:rw_math_7bm}
    \end{subfigure}
    \caption{EMPG Learning Curves (Base: QWen2.5-Math-7B)}
    \label{fig:rw_7bm}
\end{figure}

\section{Experiments}

We evaluate EM Policy Gradient (EMPG) on mathematical reasoning tasks to answer four questions:
(1) Can EMPG effectively optimize reasoning policies through
reinforcement learning?
(2) How does EMPG compare with the state-of-the-art GRPO algorithm?
(3) Does EMPG learn more efficient reasoning behaviors?
(4) What reasoning strategies emerge during training?

We conduct experiments on the GSM8K~\cite{gsm8k} and
MATH~\cite{math} benchmarks using Qwen2.5 base models ranging from
1.5B to 14B parameters. For MATH, we construct a challenging subset of
approximately 8,000 problems with difficulty levels 3--5, referred to
as \textbf{MATH Hard}. Unless otherwise specified, experiments are
performed on four NVIDIA A100 GPUs, while the 14B model is trained on
eight A100 GPUs.

\begin{table}[t]
    \caption{Test scores of EMPG and GRPO on GSM8K using Qwen2.5 base models.}
    \label{tab:srl_grpo_gsm_scores}
    \centering
    \begin{tabular}{lcc}
        \toprule
        Base model & EMPG test score & GRPO test score \\
        \midrule
        Qwen2.5-1.5B          & \textbf{0.744} & 0.739 \\
        Qwen2.5-7B            & \textbf{0.888} & 0.867 \\
        Qwen2.5-Math-7B       & \textbf{0.918} & 0.004\textsuperscript{*} \\
        Qwen2.5-14B-Instruct  & \textbf{0.952} & 0.940 \\
        \bottomrule
    \end{tabular}

    \vspace{0.5em}
    \footnotesize
    \textsuperscript{*}GRPO failed to learn in this setting.
\end{table}

\begin{table}[t]
    \caption{Test scores of EMPG and GRPO on MATH Hard using Qwen2.5 base models.}
    \label{tab:srl_grpo_math_scores}
    \centering
    \begin{tabular}{lcc}
        \toprule
        Base model & EMPG test score & GRPO test score \\
        \midrule
        Qwen2.5-1.5B          & \textbf{0.575} & 0.568 \\
        Qwen2.5-7B            & \textbf{0.742} & 0.730 \\
        Qwen2.5-Math-7B       & \textbf{0.794} & 0.755 \\
        Qwen2.5-14B-Instruct  & 0.808 & \textbf{0.816} \\
        \bottomrule
    \end{tabular}
\end{table}

\subsection{Reference Model for Fine-Tuning Reasoning Models}

In reinforcement learning from human feedback
(RLHF)~\cite{rlhf}, policy optimization is typically regularized toward
a reference model to prevent excessive policy drift. Following this
practice, we incorporate a KL regularization term into the EMPG
objective:
\begin{align*}
L(\theta)
=
\mathbb{E}_{\tau\sim D}
\!\left[
R(\tau)\log P_\theta(\tau)
\right]
-
\beta
D_{\mathrm{KL}}
\!\left(
P_\theta(\tau)
\,\|\, 
P_{\mathrm{ref}}(\tau)
\right),
\end{align*}
where $\beta$ controls the strength of regularization toward the
reference model.

\begin{figure}[t]
    \centering

    \begin{subfigure}{0.45\linewidth}
        \centering
        \includegraphics[width=\linewidth]{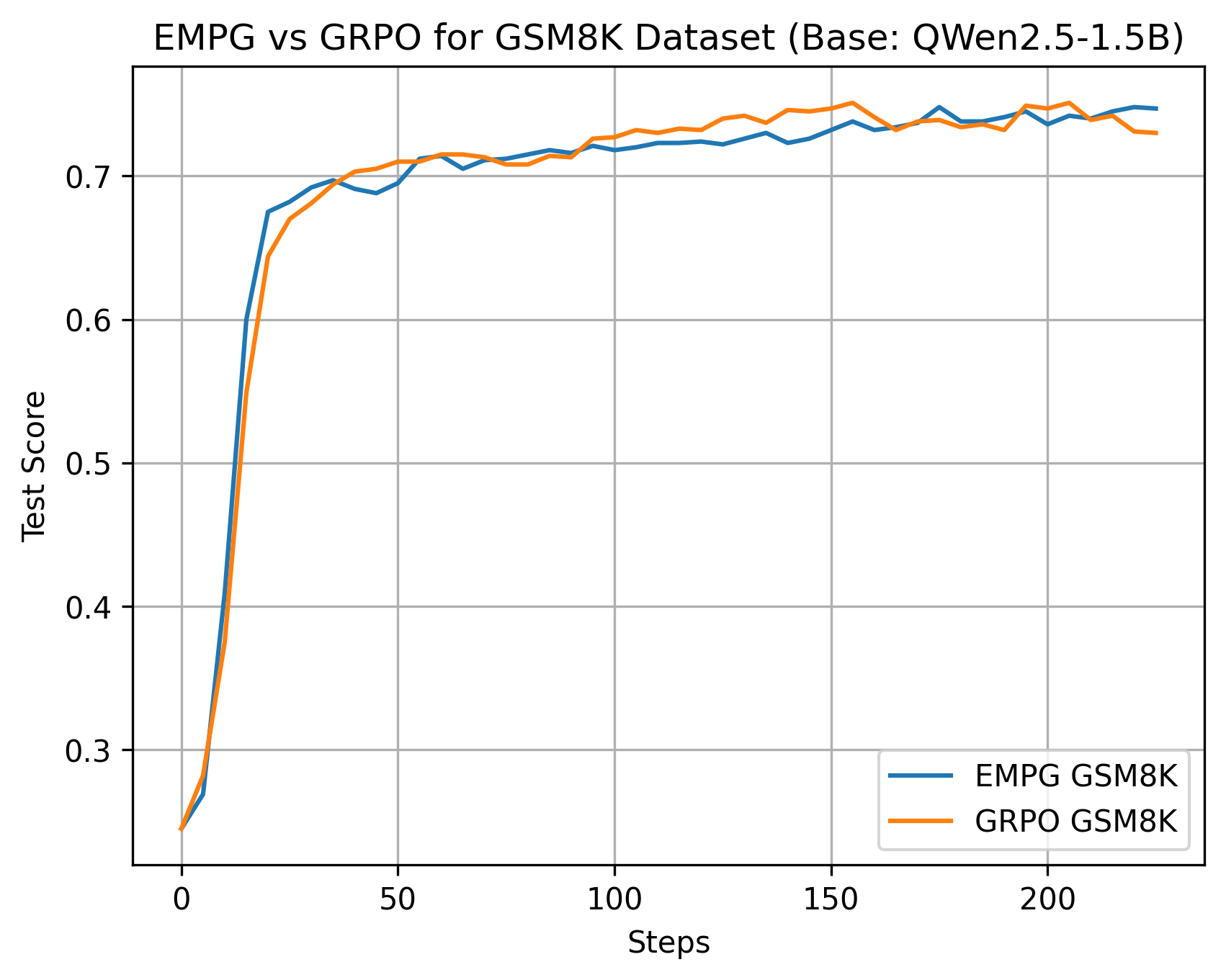}
        \caption{GSM8K.}
        \label{fig:gsm_1_5}
    \end{subfigure}
    \hfill
    \begin{subfigure}{0.45\linewidth}
        \centering
        \includegraphics[width=\linewidth]{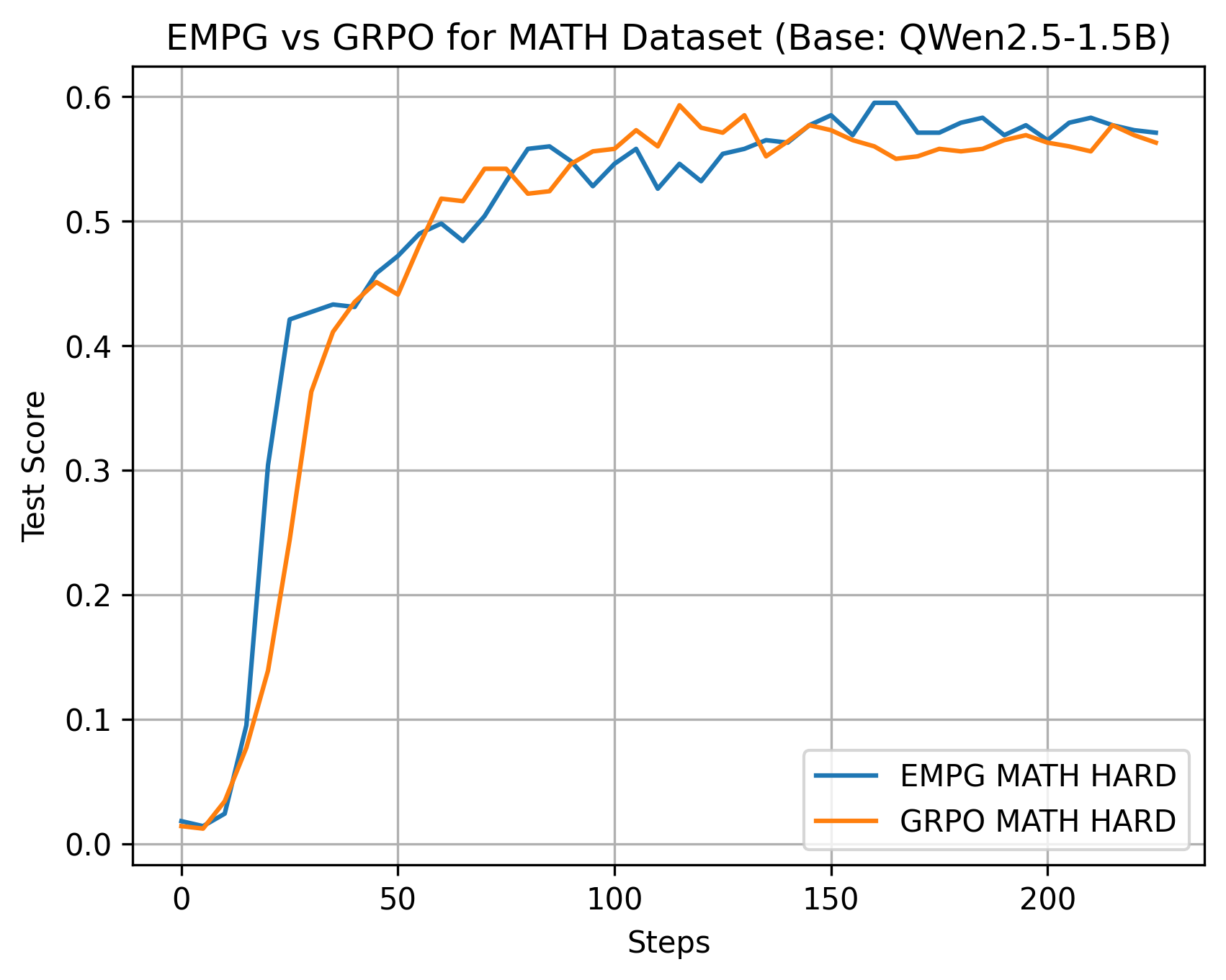}
        \caption{MATH Hard.}
        \label{fig:math_1_5}
    \end{subfigure}

    \caption{Comparison of EMPG and GRPO using Qwen2.5-1.5B as the base model on the GSM8K and MATH Hard benchmarks.}
    \label{fig:score_1_5b}
\end{figure}

Interestingly, the effectiveness of KL regularization depends on the
strength of the base model. For weaker models such as
Qwen2.5-1.5B, we find that setting $\beta=0$ yields strong performance
on both GSM8K and MATH Hard (Fig.~\ref{fig:score_1_5b}). In this
regime, removing the reference model simplifies training and reduces
memory consumption without sacrificing reasoning performance.

\begin{figure}[t]
    \centering

    \begin{subfigure}{0.45\linewidth}
        \centering
        \includegraphics[width=\linewidth]{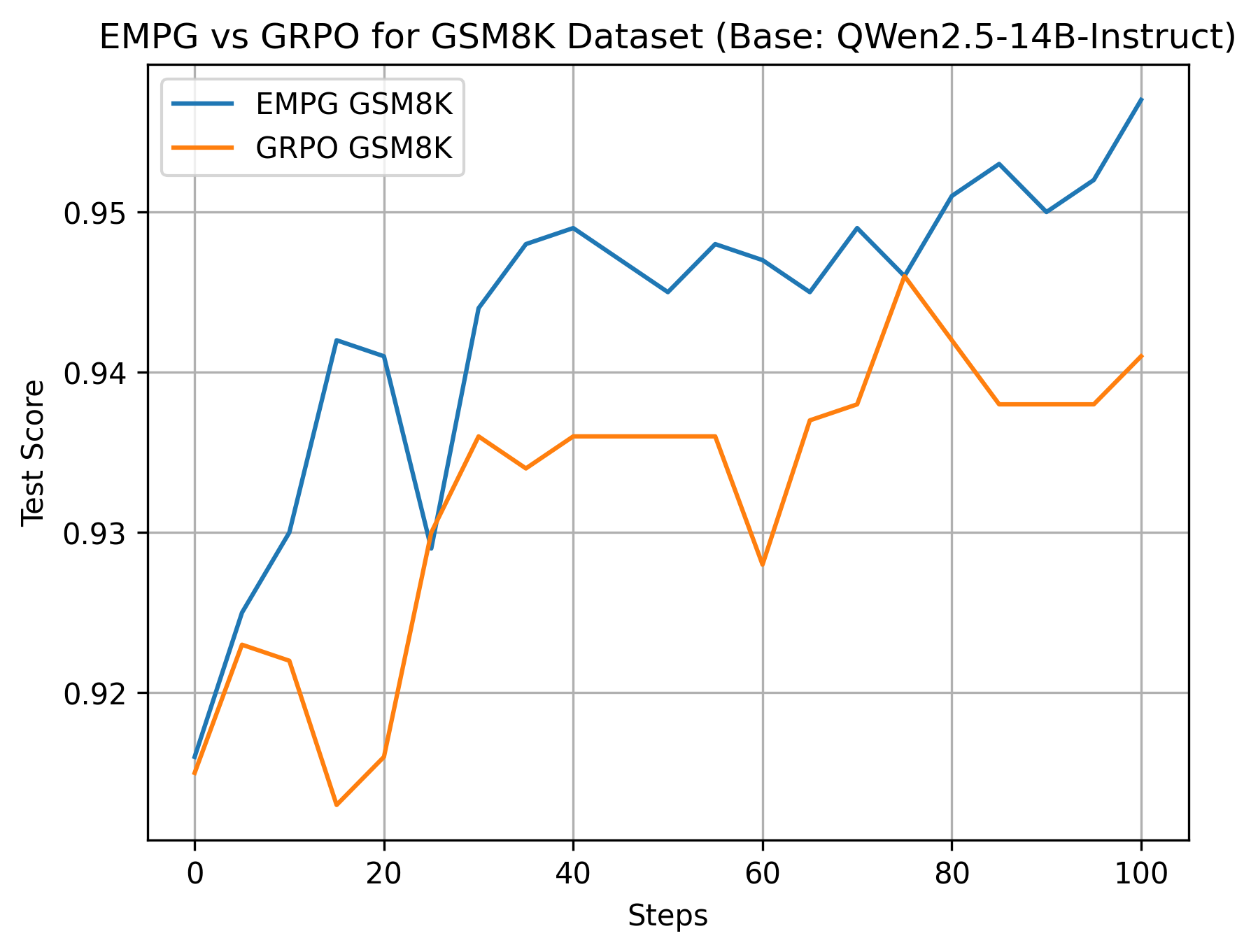}
        \caption{GSM8K.}
        \label{fig:gsm_14}
    \end{subfigure}
    \hfill
    \begin{subfigure}{0.45\linewidth}
        \centering
        \includegraphics[width=\linewidth]{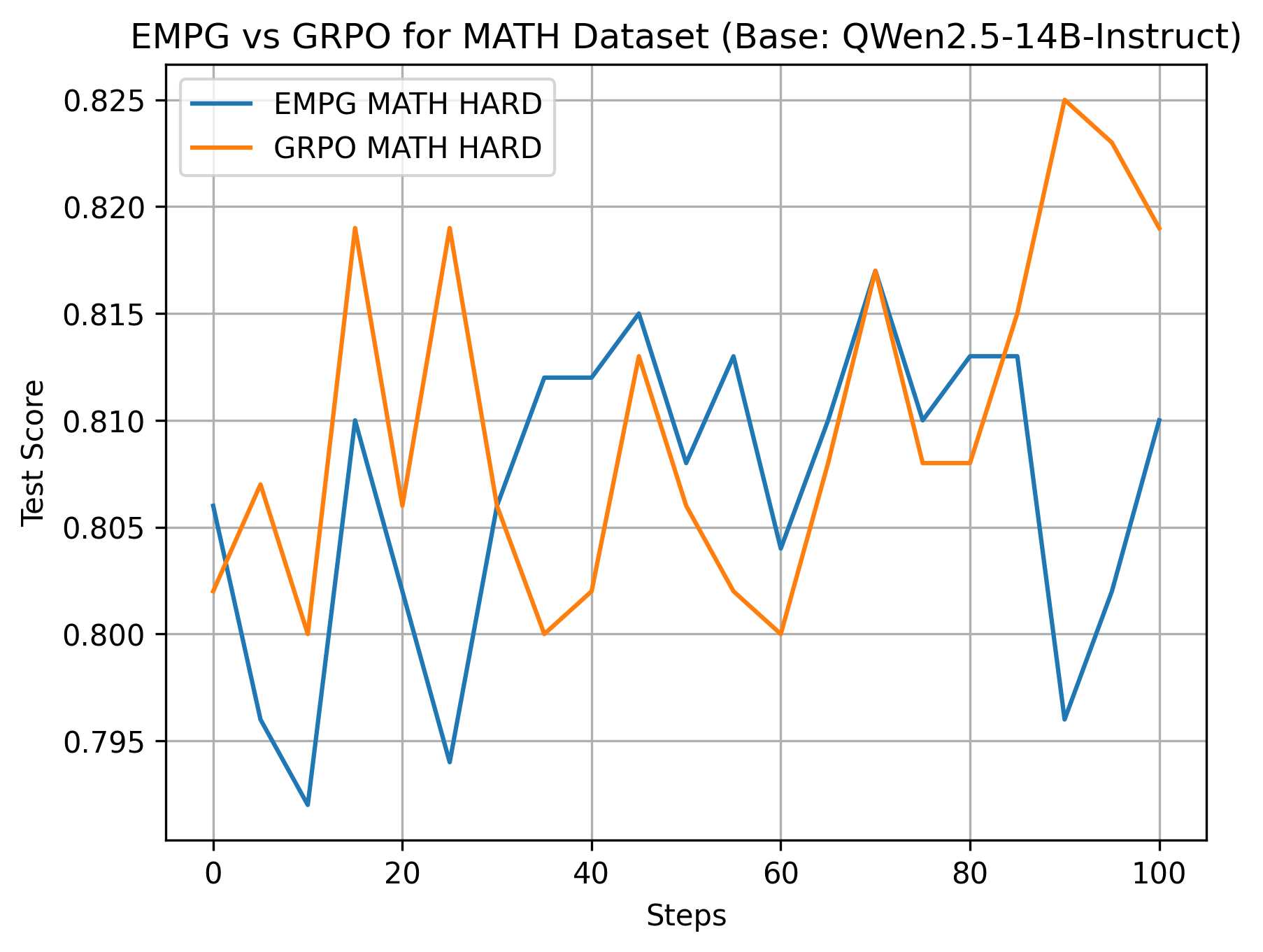}
        \caption{MATH Hard.}
        \label{fig:math_14}
    \end{subfigure}

    \caption{Comparison of EMPG and GRPO using Qwen2.5-14B-Instruct as the base model on the GSM8K and MATH Hard benchmarks.}
    \label{fig:score_14}
\end{figure}

In contrast, stronger base models such as
Qwen2.5-14B-Instruct benefit from KL regularization
(Fig.~\ref{fig:score_14}). Since these models already possess strong
reasoning capabilities, regularizing toward the reference policy helps
preserve useful behaviors while improving training stability.

\begin{figure}[t]
    \centering

    \begin{subfigure}{0.45\linewidth}
        \centering
        \includegraphics[width=\linewidth]{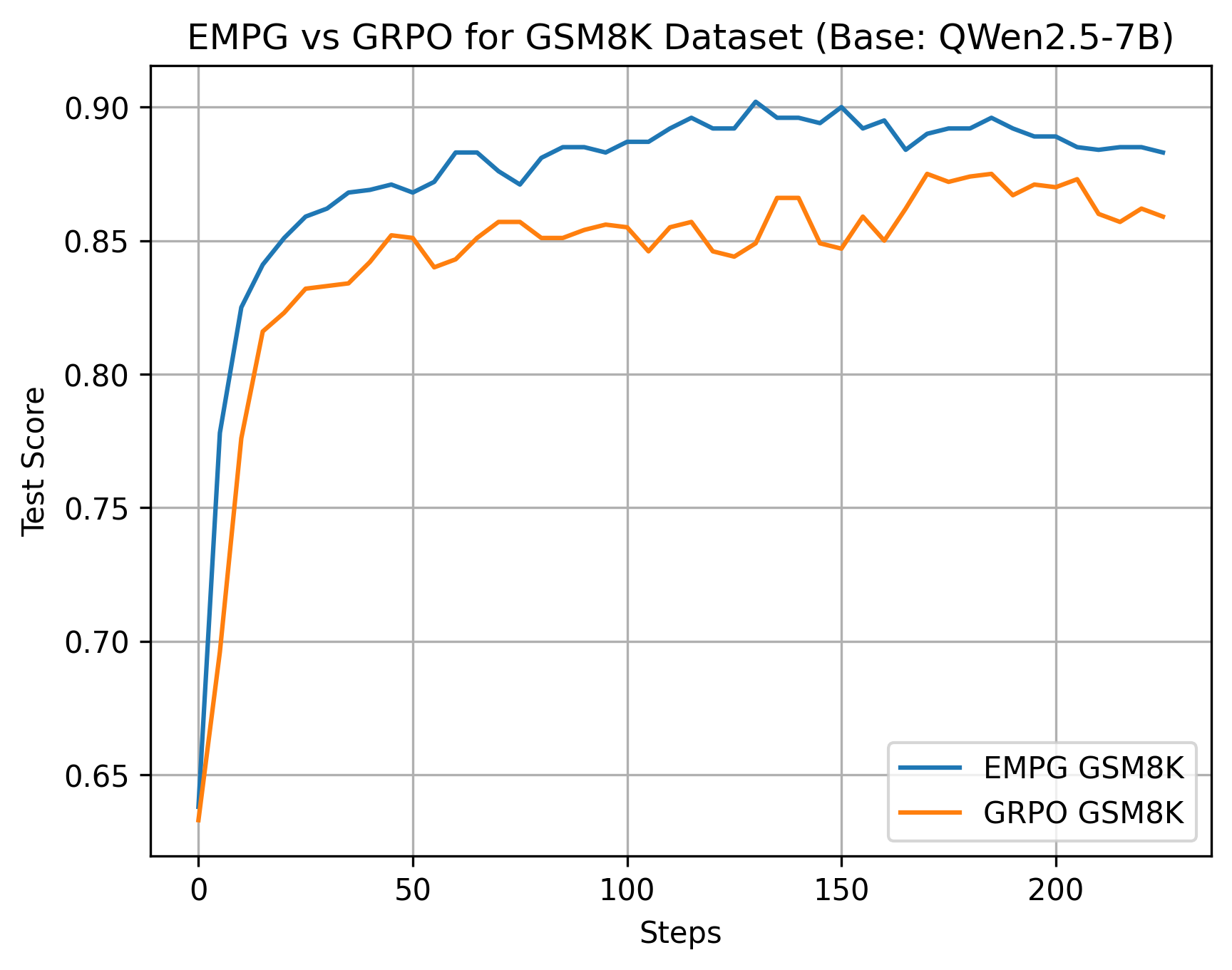}
        \caption{GSM8K.}
        \label{fig:gsm_7}
    \end{subfigure}
    \hfill
    \begin{subfigure}{0.45\linewidth}
        \centering
        \includegraphics[width=\linewidth]{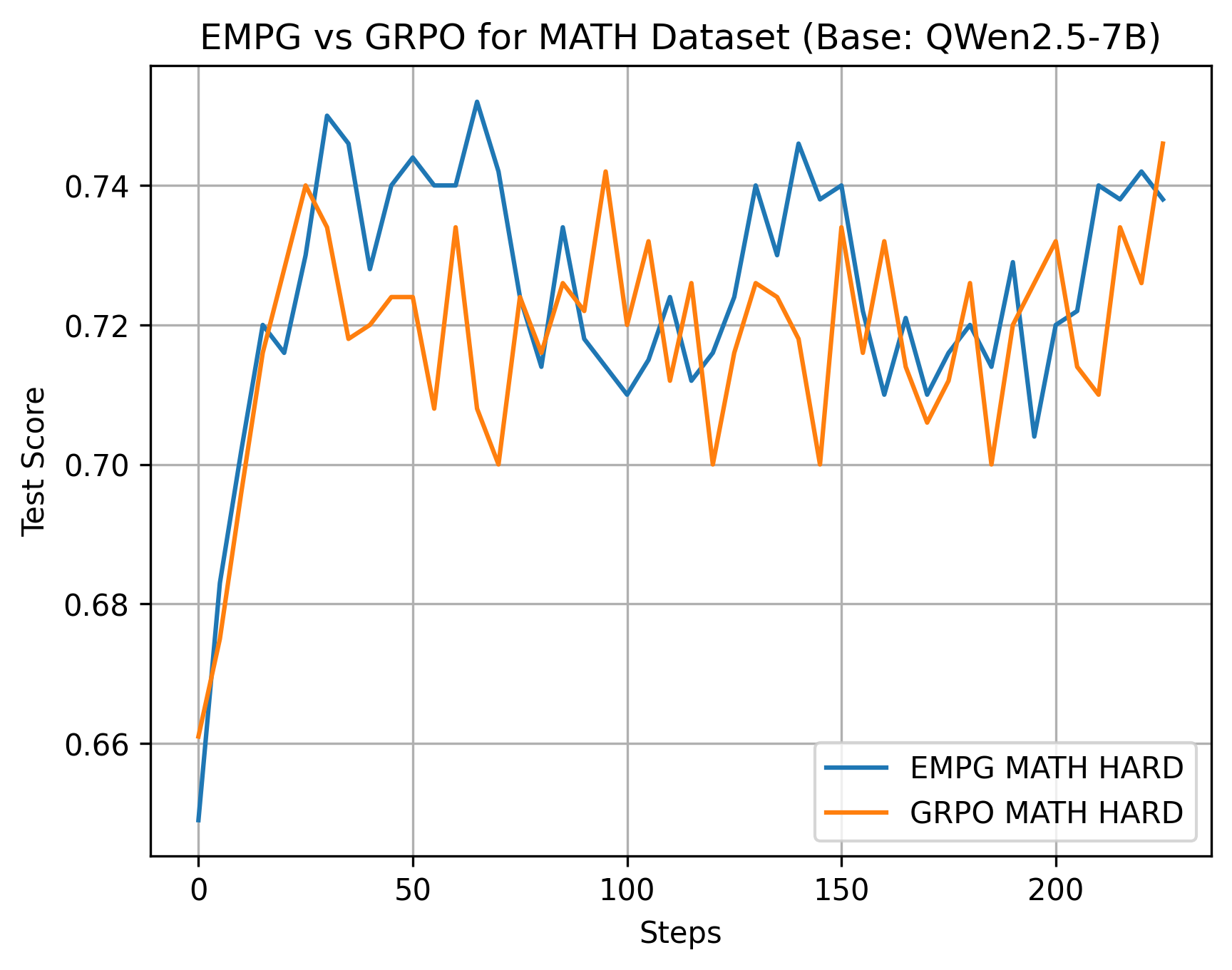}
        \caption{MATH Hard.}
        \label{fig:math_7}
    \end{subfigure}

    \caption{Comparison of EMPG and GRPO using Qwen2.5-7B as the base model on the GSM8K and MATH Hard benchmarks.}
    \label{fig:score_7b}
\end{figure}

\begin{figure}[t]
    \centering

    \begin{subfigure}{0.45\linewidth}
        \centering
        \includegraphics[width=\linewidth]{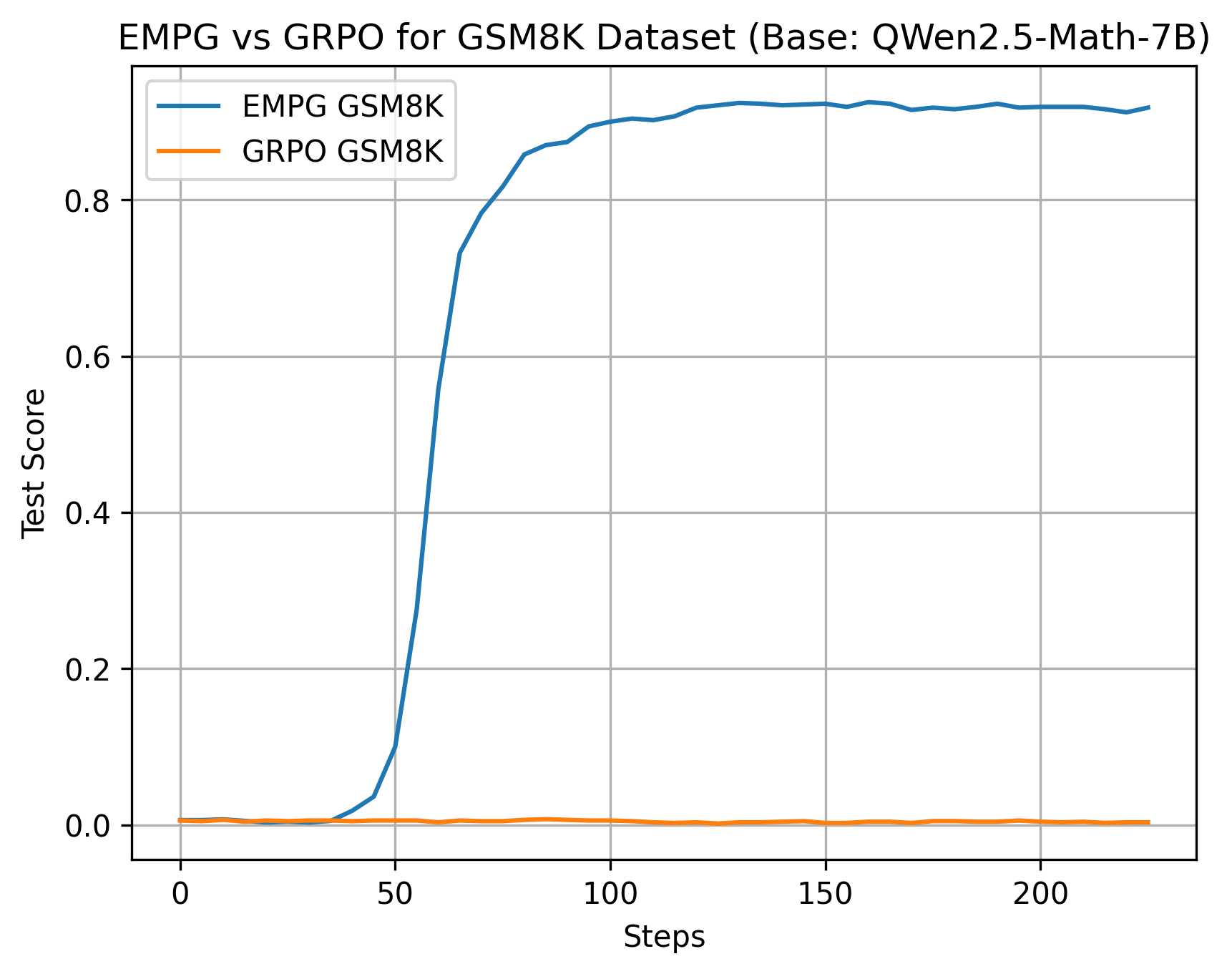}
        \caption{GSM8K.}
        \label{fig:gsm_7m}
    \end{subfigure}
    \hfill
    \begin{subfigure}{0.45\linewidth}
        \centering
        \includegraphics[width=\linewidth]{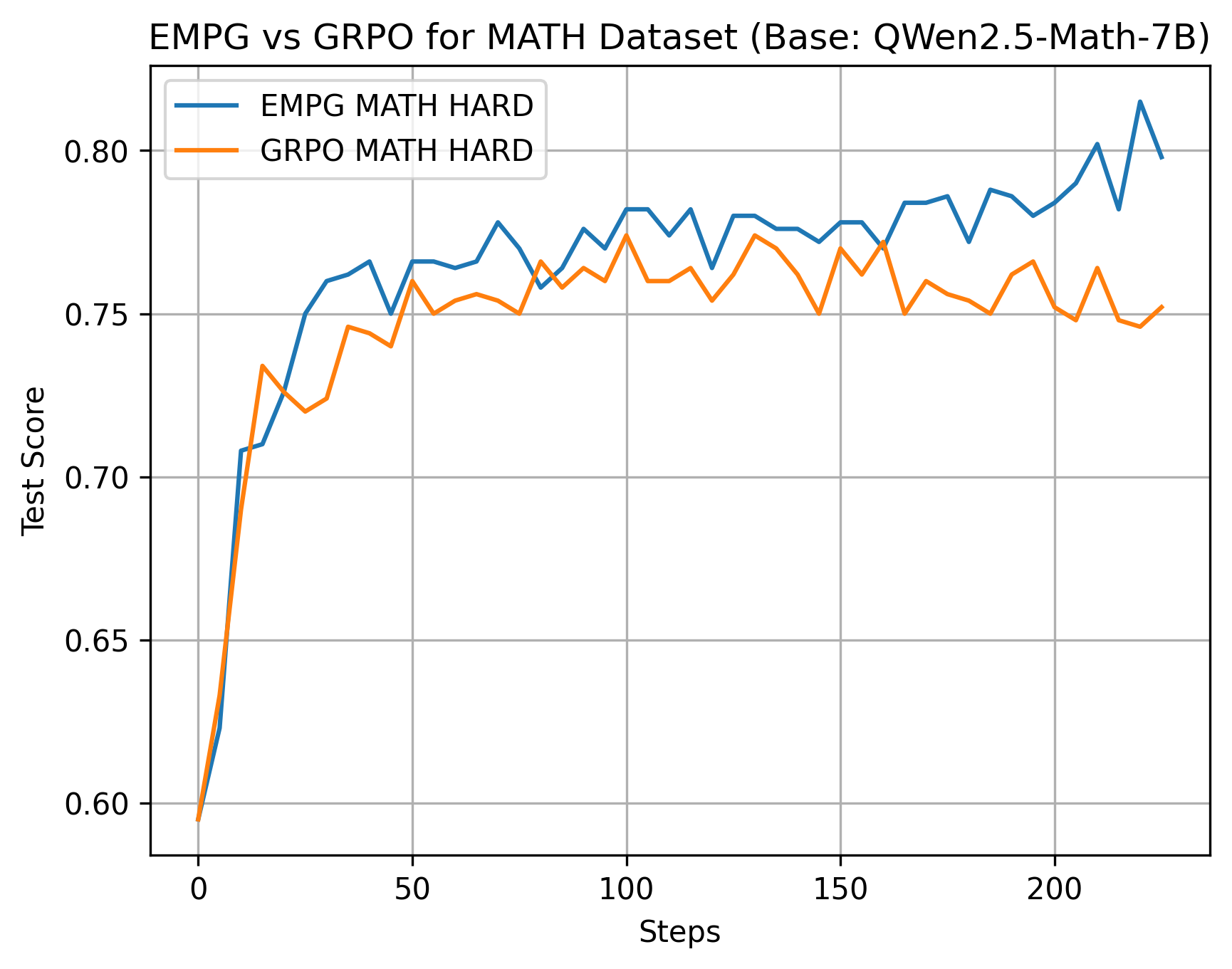}
        \caption{MATH Hard.}
        \label{fig:math_7m}
    \end{subfigure}

    \caption{Comparison of EMPG and GRPO using Qwen2.5-Math-7B as the base model on the GSM8K and MATH Hard benchmarks.}
    \label{fig:score_7mb}
\end{figure}

\subsection{Learning from Different Base Models}

We compare EMPG with GRPO across the Qwen2.5 family of base models,
including 1.5B, 7B, Math-7B, and 14B-Instruct. Unless otherwise
specified, both methods use the same training configuration following
the VERL framework, including learning rate, batch size, and optimizer
settings. Dataset-specific prompt and response lengths are adjusted only
to satisfy memory constraints. Tables~\ref{tab:srl_grpo_gsm_scores}
and~\ref{tab:srl_grpo_math_scores} report the average evaluation score
over the final ten checkpoints.

\paragraph{Weak base models.}
EMPG successfully learns effective reasoning policies even when starting
from relatively weak models such as Qwen2.5-1.5B
(Fig.~\ref{fig:score_1_5b}). Although the initial reasoning trajectories
are of low quality, EMPG rapidly improves performance through
reward-guided exploration, with most gains occurring within the first
50 training steps. On both GSM8K and MATH Hard, EMPG achieves
performance comparable to GRPO (Tables~\ref{tab:srl_grpo_gsm_scores}
and~\ref{tab:srl_grpo_math_scores}), demonstrating that the proposed
off-policy optimization remains effective even with weak initialization.

\paragraph{Medium-sized base models.}
For the 7B and Math-7B models, EMPG consistently matches or outperforms
GRPO across both benchmarks (Figs.~\ref{fig:score_7b}
and~\ref{fig:score_7mb}). On Qwen2.5-7B, EMPG achieves scores of 0.888
on GSM8K and 0.742 on MATH Hard, compared with 0.867 and 0.730 for
GRPO. The advantage is more pronounced for Qwen2.5-Math-7B, where EMPG
achieves 0.918 on GSM8K and 0.794 on MATH Hard, while GRPO fails to
improve reliably under the same training configuration
(Fig.~\ref{fig:gsm_7m}). These results suggest that EMPG can better
exploit the reasoning capabilities of stronger pretrained models.

\paragraph{Strong base models.}
With the strong Qwen2.5-14B-Instruct base model, both EMPG and GRPO
start from high initial performance and exhibit relatively modest gains
during reinforcement learning (Fig.~\ref{fig:score_14}). This behavior
is expected, as KL regularization encourages the fine-tuned policy to
remain close to an already capable reference model, reducing the need
for aggressive exploration. Nevertheless, EMPG remains competitive,
achieving a higher GSM8K score (0.952 vs.~0.940) and comparable
performance on MATH Hard (0.808 vs.~0.816).

\subsection{Reasoning Length: Shorter Can Be Better}

Recent reasoning models often improve performance by generating longer
chains of thought (CoTs), effectively allocating additional computation
during inference. For example, DeepSeek-R1~\cite{r1} exhibits steadily
increasing reasoning lengths throughout reinforcement learning
training. However, longer reasoning traces do not necessarily translate
into higher reasoning accuracy and may instead lead to
\emph{overthinking}, where additional tokens contribute little to the
final solution~\cite{cot}. Since inference latency grows with the number
of generated tokens, an effective reasoning model should ideally achieve
high accuracy with concise reasoning trajectories.

\begin{figure}[t]
    \centering

    \begin{subfigure}{0.48\linewidth}
        \centering
        \includegraphics[width=\linewidth]{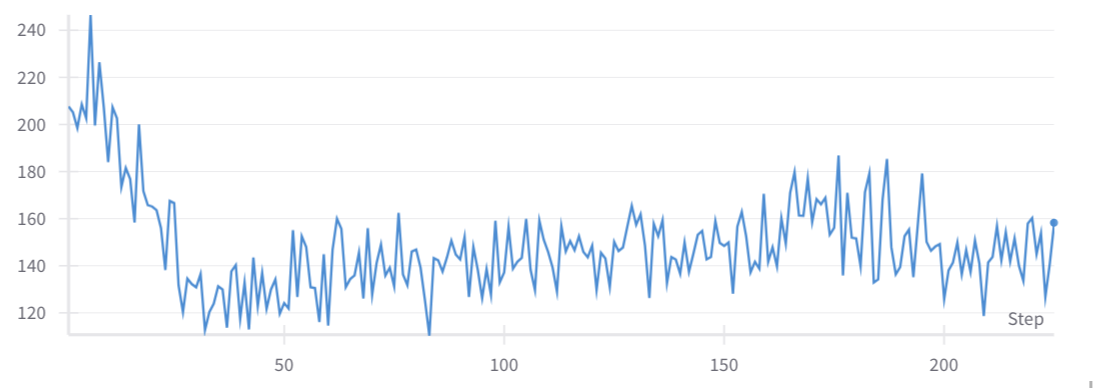}
        \caption{EMPG}
        \label{fig:srl_len_1_5}
    \end{subfigure}
    \hfill
    \begin{subfigure}{0.48\linewidth}
        \centering
        \includegraphics[width=\linewidth]{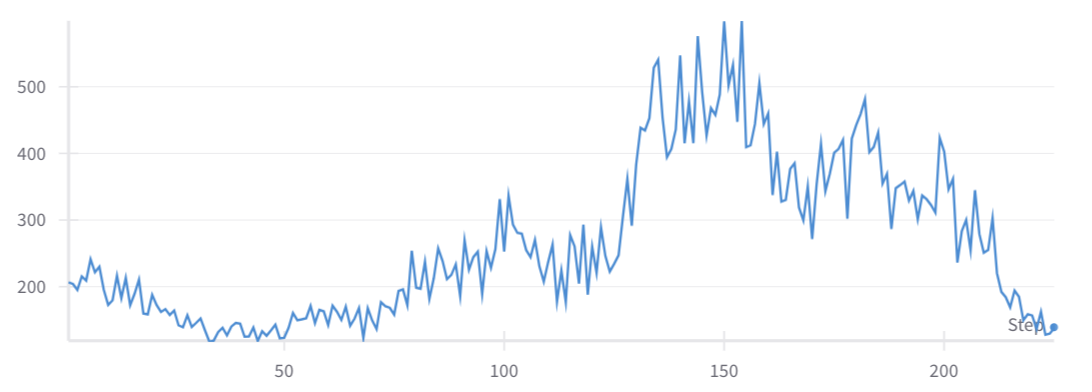}
        \caption{GRPO}
        \label{fig:grpo_len_1_5}
    \end{subfigure}

    \caption{Reasoning length comparison between EMPG and GRPO on GSM8K using Qwen2.5-1.5B.}
    \label{fig:len_1_5}
\end{figure}

\begin{figure}[t]
    \centering

    \begin{subfigure}{0.48\linewidth}
        \centering
        \includegraphics[width=\linewidth]{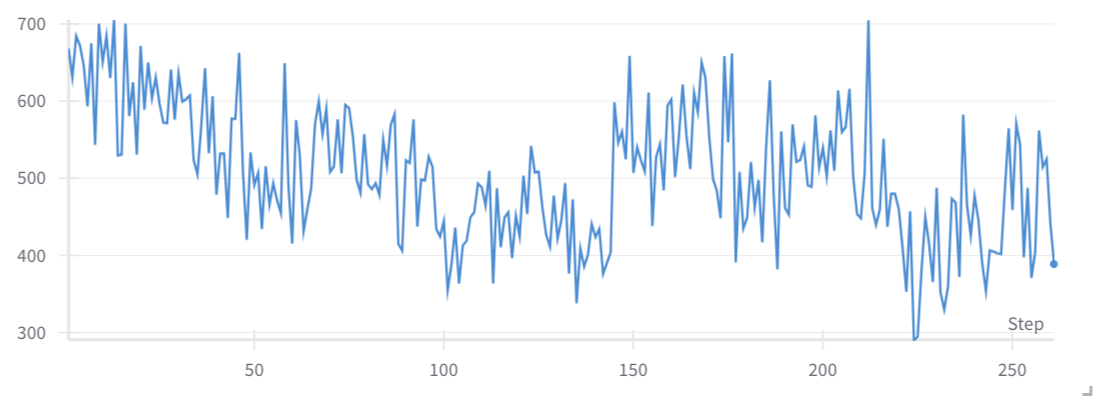}
        \caption{EMPG}
        \label{fig:srl_len_7b}
    \end{subfigure}
    \hfill
    \begin{subfigure}{0.48\linewidth}
        \centering
        \includegraphics[width=\linewidth]{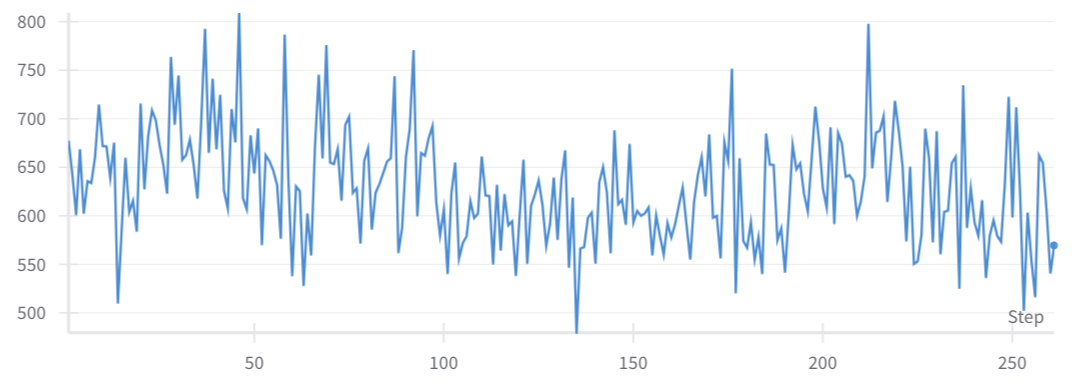}
        \caption{GRPO}
        \label{fig:grpo_len_7b}
    \end{subfigure}

    \caption{Reasoning length comparison between EMPG and GRPO on MATH using Qwen2.5-7B.}
    \label{fig:len_7b}
\end{figure}

\begin{figure}[t]
    \centering

    \begin{subfigure}{0.48\linewidth}
        \centering
        \includegraphics[width=\linewidth]{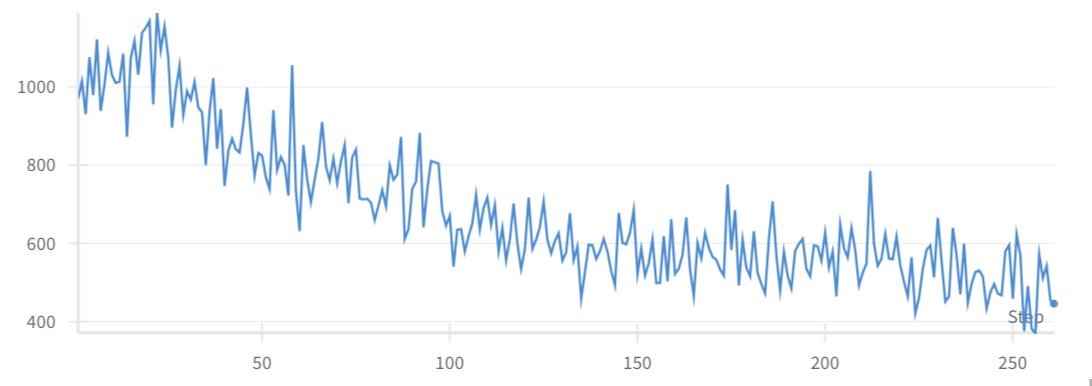}
        \caption{EMPG}
        \label{fig:srl_len_7bm}
    \end{subfigure}
    \hfill
    \begin{subfigure}{0.48\linewidth}
        \centering
        \includegraphics[width=\linewidth]{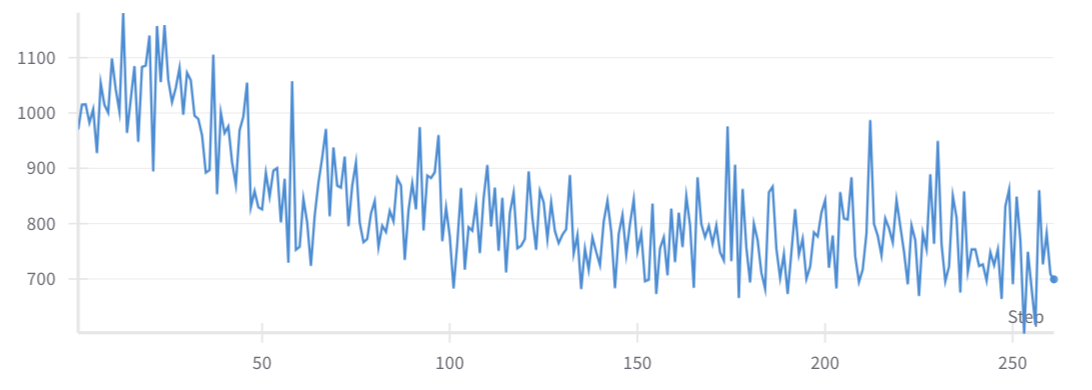}
        \caption{GRPO}
        \label{fig:grpo_len_7bm}
    \end{subfigure}

    \caption{Reasoning length comparison between EMPG and GRPO on MATH using the Qwen2.5-Math-7B base model.}
    \label{fig:len_7bm_math}
\end{figure}

\begin{figure}[t]
    \centering

    \begin{subfigure}{0.48\linewidth}
        \centering
        \includegraphics[width=\linewidth]{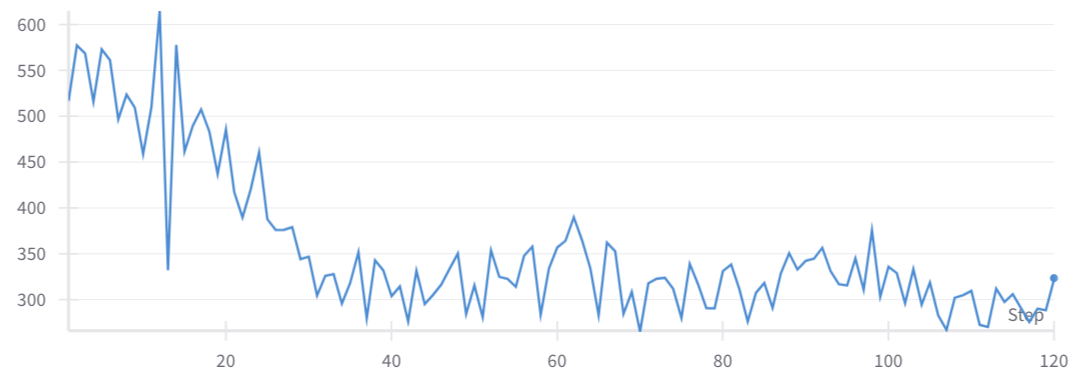}
        \caption{EMPG}
        \label{fig:srl_len_14b}
    \end{subfigure}
    \hfill
    \begin{subfigure}{0.48\linewidth}
        \centering
        \includegraphics[width=\linewidth]{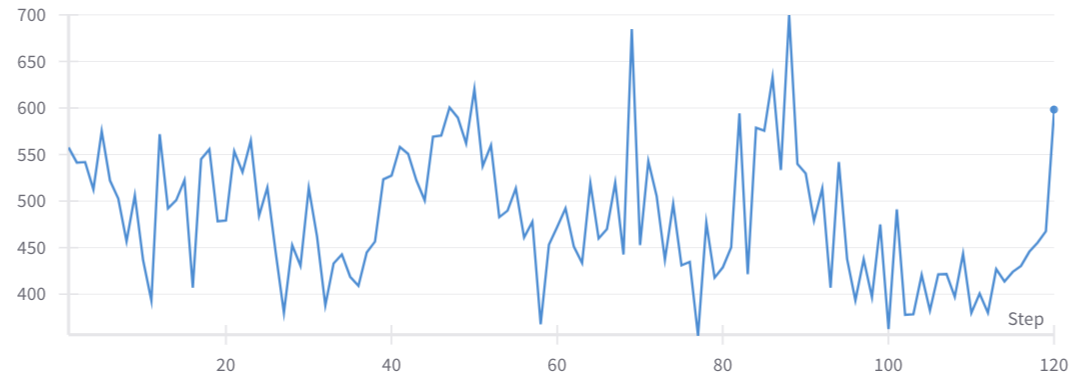}
        \caption{GRPO}
        \label{fig:grpo_len_14b}
    \end{subfigure}

    \caption{Reasoning length comparison between EMPG and GRPO on GSM8K using the Qwen2.5-14B-Instruct base model.}
    \label{fig:len_14b}
\end{figure}

We therefore compare the response lengths produced by EMPG and GRPO
across different Qwen2.5 base models
(Figs.~\ref{fig:len_1_5}--\ref{fig:len_14b}). Despite achieving
comparable or better reasoning performance, EMPG consistently generates
shorter reasoning trajectories than GRPO. For example, with the
Qwen2.5-1.5B model on GSM8K, EMPG converges to an average response
length of approximately 120--180 tokens, whereas GRPO typically
generates 200--500 tokens
(Fig.~\ref{fig:len_1_5}). On the more challenging MATH Hard benchmark
using Qwen2.5-Math-7B, both methods produce longer reasoning traces, but
EMPG stabilizes around 400--600 tokens, compared with roughly 700--900
tokens for GRPO (Fig.~\ref{fig:len_7bm_math}).

These results suggest that EMPG learns more efficient reasoning
policies, achieving competitive reasoning accuracy while requiring
substantially fewer generated tokens. This reduction in reasoning length
can directly improve inference efficiency by lowering decoding latency
and computational cost.

\begin{comment}
\begin{figure}[htbp]
    \centering
    \begin{subfigure}[b]{0.48\textwidth}
        \includegraphics[width=\textwidth]{reasoning_len/em_math_7b_gsm_len.png}
        \caption{EMPG}
        \label{fig:srl_gsm_len_7bm}
    \end{subfigure}
    \hfill
    \begin{subfigure}[b]{0.48\textwidth}
        \includegraphics[width=\textwidth]{reasoning_len/grpo_7bmath_gsm_len.png}
        \caption{GRPO}
        \label{fig:grpo_gsm_len_7bm}
    \end{subfigure}
    \caption{EMPG vs GRPO Reasoning Length (GSM8K, Base: QWen2.5-Math-7B)}
    \label{fig:len_7bm}
\end{figure}
\end{comment}

\subsection{Case Study: Emergent Reasoning Behaviors}

Beyond benchmark accuracy, we examine the reasoning trajectories
generated by EMPG to better understand the behaviors acquired through
reinforcement learning. The fine-tuned models produce explicit
chain-of-thought (CoT) reasoning together with the final answer,
revealing a variety of structured reasoning strategies.

Across the examples in
Figs.~\ref{fig:math_example_1}--\ref{fig:math_example_5}, we observe
several recurring behaviors. The models decompose complex mathematical
problems into simpler subproblems and perform step-by-step symbolic
reasoning (Figs.~\ref{fig:math_example_1}
and~\ref{fig:math_example_3}). They frequently verify intermediate
results by generating executable Python code or by checking solutions
through alternative derivations
(Figs.~\ref{fig:math_example_1}--\ref{fig:math_example_3}). More
interestingly, the models exhibit self-reflection and backtracking:
after detecting inconsistencies in an initial solution, they revisit
earlier reasoning steps, correct errors, and pursue alternative
strategies before producing the final answer
(Figs.~\ref{fig:math_example_4}
and~\ref{fig:math_example_5}).

These examples suggest that EMPG learns more than simply generating
longer reasoning traces. Instead, reinforcement learning encourages
structured reasoning behaviors, including problem decomposition,
step-by-step deduction, self-verification, error correction, and
backtracking. Such behaviors closely resemble strategies commonly
employed by human problem solvers and provide qualitative evidence for
the effectiveness of the proposed framework.

\section{Related Work}
\label{sec:related_work}

\paragraph{Learning to reason and test-time scaling.}
Improving the reasoning capabilities of large language models has become
a central research direction. Chain-of-Thought (CoT)
prompting~\cite{cotw} demonstrates that generating intermediate
reasoning steps substantially improves performance on complex reasoning
tasks. Building on this idea, Tree of Thoughts
(ToT)~\cite{tot} explores multiple reasoning branches through explicit
search, while Self-Taught Reasoner (STaR)~\cite{star} iteratively
improves reasoning by fine-tuning on successfully generated reasoning
traces. A complementary line of work improves reasoning by allocating
additional computation during inference. Verifier-guided test-time
scaling~\cite{tts} improves performance by sampling and selecting among
multiple candidate solutions, while recent systems such as
T1~\cite{t1} further investigate how reasoning scales with extended
inference, iterative refinement, and self-verification. Collectively,
these approaches demonstrate that richer reasoning trajectories and
greater inference-time computation are both effective mechanisms for
improving reasoning performance.

\paragraph{Reinforcement learning for reasoning.}
Recent reasoning models, including OpenAI's
O1~\cite{o1}, DeepSeek-R1~\cite{r1}, Kimi
k1.5~\cite{kimi_k15}, and QwQ-32B~\cite{qwen}, have demonstrated that
large-scale reinforcement learning can substantially improve
mathematical reasoning, code generation, and general problem-solving
capabilities. These advances build upon decades of research in policy
optimization, from REINFORCE~\cite{reinforce} and
TRPO~\cite{trpo} to PPO~\cite{ppo}. More recently, Group Relative Policy
Optimization (GRPO)~\cite{grpo} has become a widely adopted algorithm
for post-training reasoning language models.

Recent work has also explored reinforcement learning with verifiable
rewards (RLVR), where correctness can be automatically evaluated for
domains such as mathematics and code generation. Complementary studies
highlight the importance of exploration, long-context reasoning, and
stable optimization for effectively training reasoning language
models~\cite{kimi_k15,t1}.

Our work differs from the above approaches in perspective rather than in
architecture or search strategy. Instead of modifying existing
reinforcement learning objectives, we formulate learning to reason as an
Expectation-Maximization (EM) optimization problem over latent
reasoning trajectories, providing a new optimization framework for
reasoning language models.

\section{Conclusion}

We introduced \textbf{EM Policy Gradient (EMPG)}, a simple and
principled reinforcement learning framework for training reasoning
language models. Rather than viewing reinforcement learning over
reasoning trajectories as an on-policy optimization problem, we
formulated it as an Expectation-Maximization (EM) optimization problem.
This perspective naturally separates trajectory generation from policy
optimization, leading to a practical off-policy policy-gradient
algorithm and establishing a monotonic improvement guarantee under exact
E- and improving M-steps.

The resulting algorithm directly optimizes the EM surrogate objective
without trajectory-level importance weighting or likelihood-ratio
clipping, yielding a substantially simpler optimization procedure than
existing methods such as PPO~\cite{ppo} and GRPO~\cite{grpo}. Empirical
results on GSM8K and MATH Hard demonstrate that EMPG achieves
performance comparable to or better than GRPO across multiple Qwen2.5
base models while producing concise reasoning trajectories and rich
reasoning behaviors, including subproblem decomposition,
self-verification, reflection, and backtracking.

More broadly, this work suggests that reinforcement learning for
reasoning can be viewed through the lens of latent-variable optimization
rather than solely as policy optimization. We hope that the proposed EM
formulation provides a foundation for developing new optimization
algorithms, richer variational objectives, and more effective
reinforcement learning methods for reasoning language models.

\section*{Acknowledgments}

We thank Professor Qiang Liu of the University of Texas at Austin for valuable feedback and insightful discussions. We are also grateful to Jonah Philion for thoughtful discussions and constructive suggestions that helped improve this paper.

\clearpage
\section{Supplementary Material}
\label{sec:supplementary}

\subsection{Proof of Theorem~\ref{thm:monotonic_improvement}}
\label{sec:supplementary_monotonic_proof}

Here, we establish a monotonic-improvement guarantee for EM Policy
Gradient.

\begin{proof}

Recall that
\[
J(\theta)
:=
\sum_{\tau}
P_\theta(\tau)R(\tau).
\]

For any variational trajectory distribution \(q(\tau)\)
\begin{align*} \log J(\theta) &= \log \sum_{\tau} q(\tau) \frac{ P_\theta(\tau)R(\tau) }{ q(\tau) }. \end{align*}

Define the posterior variational distribution
\[
q_\theta^\star(\tau)
:=
\frac{
P_\theta(\tau)R(\tau)
}{
J(\theta)
}.
\]

Then
\begin{align*}
\mathbb{D}_{\mathrm{KL}}
\left(
q
\,\|\, 
q_\theta^\star
\right)
&:=
\sum_{\tau}
q(\tau)
\log
\frac{
q(\tau)
}{
q_\theta^\star(\tau)
}
=
\sum_{\tau}
q(\tau)
\log
\frac{
q(\tau)J(\theta)
}{
P_\theta(\tau)R(\tau)
}
=
\log J(\theta)
-
\mathcal{L}(q,\theta).
\end{align*}

Therefore
\begin{align*}
\label{eq:elbo_decomposition}
\log J(\theta)
=
\mathcal{L}(q,\theta)
+
\mathbb{D}_{\mathrm{KL}}
\left(
q
\,\|\, 
q_\theta^\star
\right).
\end{align*}

Since the KL divergence is nonnegative
\[
\log J(\theta)
\ge
\mathcal{L}(q,\theta).
\]

At iteration \(t\), the exact E-step sets
\[
q_t(\tau)
=
q_{\theta_t}^\star(\tau)
=
\frac{
P_{\theta_t}(\tau)R(\tau)
}{
J(\theta_t)
}.
\]

Hence
\[
\mathbb{D}_{\mathrm{KL}}
\left(
q_t
\,\|\, 
q_{\theta_t}^\star
\right)
=
0,
\]

so the lower bound is tight
\[
\log J(\theta_t)
=
\mathcal{L}(q_t,\theta_t).
\]

By assumption, the M-step increases the lower bound
\[
\mathcal{L}(q_t,\theta_{t+1})
\ge
\mathcal{L}(q_t,\theta_t).
\]

Applying the lower-bound property at \(\theta_{t+1}\) yields
\[
\log J(\theta_{t+1})
\ge
\mathcal{L}(q_t,\theta_{t+1}).
\]

Combining the preceding relations gives
\[
\log J(\theta_{t+1})
\ge
\mathcal{L}(q_t,\theta_{t+1})
\ge
\mathcal{L}(q_t,\theta_t)
=
\log J(\theta_t).
\]

Because the logarithm is strictly increasing, we have
\[
J(\theta_{t+1})
\ge
J(\theta_t).
\]
\end{proof}

\clearpage
\subsection{Qualitative Reasoning Examples}

% ============================================================
% Example 1
% ============================================================

\begin{reasonbox}

\textbf{Problem}

The equation \(x^2-6x+2=29\) has two solutions, \(a\) and \(b\), with
\(a\geq b\). What is the value of \(2a-3b\)?

\smallskip

\textbf{COT and Response:}

To solve the quadratic equation \(x^2-6x+2=29\),
\textbf{we can follow these steps:}

1. Rearrange the equation to standard quadratic form:
\(x^2-6x+2-29=0\), which simplifies to
\(x^2-6x-27=0\).

2. Use the quadratic formula
\[
    x=\frac{-b\pm\sqrt{b^2-4ac}}{2a}
\]
to find the solutions, where \(a=1\), \(b=-6\), and \(c=-27\).

3. Calculate the discriminant:
\[
    \Delta
    =
    b^2-4ac
    =
    (-6)^2-4(1)(-27)
    =
    36+108
    =
    144.
\]

4. Use the quadratic formula to find the solutions:
\[
    x
    =
    \frac{6\pm\sqrt{144}}{2\cdot1}
    =
    \frac{6\pm12}{2}.
\]

5. This gives us the solutions
\(x=\frac{6+12}{2}=9\) and
\(x=\frac{6-12}{2}=-3\).
So, \(a=9\) and \(b=-3\).

6. Finally, we calculate
\[
    2a-3b
    =
    2(9)-3(-3)
    =
    18+9
    =
    27.
\]

\textbf{Let's implement this in Python to verify the solution.}

\begin{lstlisting}[style=pythonStyle]
import math

# Coefficients of the quadratic equation
a_coeff = 1
b_coeff = -6
c_coeff = -27

# Calculate the discriminant
discriminant = b_coeff**2 - 4*a_coeff*c_coeff

# Calculate the two solutions using the quadratic formula
x1 = (-b_coeff + math.sqrt(discriminant)) / (2*a_coeff)
x2 = (-b_coeff - math.sqrt(discriminant)) / (2*a_coeff)

# Determine a and b with a >= b
a = max(x1, x2)
b = min(x1, x2)

# Calculate 2a - 3b
result = 2*a - 3*b
print(result)
\end{lstlisting}

\noindent\textbf{Output:} \texttt{27.0}

\noindent Therefore, the value of \(2a-3b\) is \(\boxed{27}\).

\end{reasonbox}

\captionof{figure}{MATH problem example: think step-by-step and
self-verification with Python code.}
\label{fig:math_example_1}

% ============================================================
% Example 2
% ============================================================

\clearpage

\begin{reasonbox}

\textbf{Problem}

Find the smallest \(n>3\) such that
\((1+2+3+\cdots+n)^2\) is a perfect fourth power.

\smallskip

\textbf{COT and Response:}

To solve the problem, we need to find the smallest integer \(n>3\)
such that \((1+2+3+\cdots+n)^2\) is a perfect fourth power. The sum
of the first \(n\) positive integers is given by
\[
    S=\frac{n(n+1)}{2}.
\]

So we need
\[
    S^2
    =
    \left(\frac{n(n+1)}{2}\right)^2
\]
to be a perfect fourth power. This means that
\(\frac{n(n+1)}{2}\) must be a perfect square. Let's denote
\[
    \frac{n(n+1)}{2}=k^2
\]
for some integer \(k\). Then, we need to find the smallest \(n>3\)
such that \(\frac{n(n+1)}{2}\) is a perfect square.

We can iterate over \(n\) starting from 4 and check if
\(\frac{n(n+1)}{2}\) is a perfect square.

\textbf{Here's the Python code to find the solution:}

\begin{lstlisting}[style=pythonStyle]
import math

def is_perfect_square(x):
    root = int(math.isqrt(x))
    return x == root * root

n = 4
while True:
    sum_of_integers = n * (n + 1) // 2
    if is_perfect_square(sum_of_integers):
        break
    n += 1
\end{lstlisting}

\noindent\textbf{Output:} \texttt{8}

The smallest \(n>3\) such that
\((1+2+3+\cdots+n)^2\) is a perfect fourth power is \(n=8\).

\textbf{To verify, let's check the sum of the first 8 positive
integers:}
\[
    S=\frac{8\cdot9}{2}=36.
\]

And \(36^2=1296\), which is \(36^2=6^4\), a perfect fourth power.

So the final answer is
\[
    \boxed{8}.
\]

\end{reasonbox}

\captionof{figure}{MATH problem example: self-verification and problem
solving with Python code.}
\label{fig:math_example_2}

% ============================================================
% Example 3
% ============================================================

\clearpage

\begin{reasonbox}

\textbf{Problem (AIME 2010)}

Let \(P(x)\) be a quadratic polynomial with real coefficients satisfying
the inequality
\[
    x^2-2x+2
    \leq
    P(x)
    \leq
    2x^2-4x+3,
\]
for all real numbers \(x\), and suppose that \(P(11)=181\).
Find \(P(16)\).

\smallskip

\textbf{COT and Response:}

To solve this problem, we need to find the quadratic polynomial
\(P(x)=ax^2+bx+c\) that satisfies the given inequalities and the
condition \(P(11)=181\). The inequalities are
\[
    x^2-2x+2
    \leq
    P(x)
    \leq
    2x^2-4x+3.
\]

First, let's consider the equality case for \(P(x)\). Since \(P(x)\)
is a quadratic polynomial, it must be a linear combination of the two
bounding polynomials. Let's assume
\[
    P(x)
    =
    k(x^2-2x+2)
    +(1-k)(2x^2-4x+3)
\]
for some constant \(k\). Simplifying this, we get
\[
\begin{aligned}
    P(x)
    &=
    kx^2-2kx+2k
    +2(1-k)x^2-4(1-k)x+3(1-k) \\
    &=
    (2-k)x^2+(2k-4)x+(3-k).
\end{aligned}
\]

We know that \(P(11)=181\), so we can substitute \(x=11\) into the
polynomial and solve for \(k\):
\[
    P(11)
    =
    (2-k)(11^2)+(2k-4)(11)+(3-k)
    =
    181,
    \qquad
    k=\frac{1}{5}.
\]

Now we can substitute \(k=\frac{1}{5}\) back into the polynomial
\(P(x)\):
\[
\begin{aligned}
    P(x)
    &=
    \left(2-\frac{1}{5}\right)x^2
    +
    \left(2\left(\frac{1}{5}\right)-4\right)x
    +
    \left(3-\frac{1}{5}\right) \\
    &=
    \frac{9}{5}x^2
    -
    \frac{18}{5}x
    +
    \frac{14}{5}.
\end{aligned}
\]

Now we need to find \(P(16)\):
\[
    P(16)
    =
    \frac{9}{5}(16^2)
    -
    \frac{18}{5}(16)
    +
    \frac{14}{5}
    =
    406.
\]

So the final answer is
\[
    \boxed{406}.
\]

\textbf{Let's verify this with Python code to ensure the calculations
are correct.}

\begin{lstlisting}[style=pythonStyle]
# Define the polynomial P(x) with the found coefficients
def P(x):
    return (9/5)*x**2 - (18/5)*x + (14/5)

# Calculate P(16)
P_16 = P(16)
print(P_16)
\end{lstlisting}

\noindent\textbf{Output:} \texttt{406}

\end{reasonbox}

\captionof{figure}{MATH problem example: step-by-step derivation and
code-based verification.}
\label{fig:math_example_3}

% ============================================================
% Example 4
% ============================================================

\clearpage

\begin{reasonbox}

\textbf{Problem (HMMT 2009):}

A torus (donut) having inner radius 2 and outer radius 4 sits on a flat
table. What is the radius of the largest spherical ball that can be
placed on top of the center torus so that the ball still touches the
horizontal plane? (skip \ldots)

\smallskip

\textbf{COT and Response:}

\textbf{Let's solve the problem step-by-step.} We'll first understand
the geometry and then use Python and \texttt{sympy} to find the radius
of the largest spherical ball that can be placed on top of the center
of the torus while still touching the horizontal plane.

1. \textbf{Understanding the Geometry:}

\quad-- The torus has an inner radius of 2 and an outer radius of 4.
The center of the tube of the torus is located at \((3,0,1)\), and the
radius of the tube is 1. The torus is formed by revolving this circle
around the \(z\)-axis. (skip \ldots)

2. \textbf{Formulating the Problem:}

\quad-- The distance from the center of the spherical ball
(at \((0,0,h)\)) to the center of the tube of the torus
(at \((3,0,1)\)) must be equal to \(r-1\), because the ball touches
the inner surface of the torus. (skip \ldots)

3. \textbf{Solving the Equation:}

\quad-- Square both sides of the equation to remove the square root:
\[
    9+(r-1)^2=(r-1)^2.
\]

\quad-- This simplifies to \(9=0\), which is \textbf{incorrect}.
We need to \textbf{re-evaluate} the distance condition.

\textbf{Let's correct and solve this step-by-step in Python using
\texttt{sympy}:} (skip \ldots)

\noindent\textbf{Output:}
\texttt{IndexError: list index out of range}

\textbf{It seems there was an issue with the previous approach. Let's
re-evaluate the problem more carefully.} The correct approach should
be to solve the equation derived from the geometry of the problem
correctly.

1. \textbf{Re-evaluate the Geometry:}

\quad-- The center of the spherical ball is at \((0,0,r)\).

\quad-- The distance from the center of the spherical ball to the center
of the tube of the torus, which is at \((3,0,1)\), is
\[
    \sqrt{3^2+(r-1)^2}
    =
    \sqrt{9+(r-1)^2}.
\]

\quad-- This distance must be equal to \(r+1\).

2. \textbf{Formulate the Equation:}

\quad-- We have
\[
    \sqrt{9+(r-1)^2}=r+1.
\]

\quad-- This simplifies to
\[
    9=4r,
    \qquad
    r=\frac{9}{4}.
\]

Let's solve this using Python and sympy:

\begin{lstlisting}[style=pythonStyle]
import sympy as sp

# Define the radius of the spherical ball
r = sp.symbols('r')

# The equation we derived
distance_eq = sp.sqrt(9 + (r-1)**2) - (r + 1)

# Solve the equation
solution = sp.solve(distance_eq, r)

# Filter out the valid solution
valid_solution = [sol.evalf() for sol in solution if sol > 0]

print(valid_solution[0])
\end{lstlisting}

\noindent\textbf{Output:} \texttt{9/4}

\textbf{Final Answer:} The radius of the largest spherical ball that
can be placed on top of the center of the torus and still touch the
horizontal plane is
\(\fbox{\(\frac{9}{4}\)}\).

\end{reasonbox}

\captionof{figure}{MATH problem example: step-by-step reasoning,
backtracking, reflection, and self-verification.}
\label{fig:math_example_4}

% ============================================================
% Example 5
% ============================================================

\clearpage

\begin{reasonbox}

\textbf{Problem}

You have seven bags of gold coins. Each bag contains the same number
of coins. One day, you find a bag containing 53 coins. You decide to
redistribute all the coins so that each of the eight bags, including
the new one, contains the same number of coins. After redistribution,
you note that the total number of coins exceeds 200. What is the
smallest number of coins you could have had in each of the original
seven bags before finding the additional bag of 53 coins?

\smallskip

\textbf{COT and Response:}

To solve this problem, let's denote the number of coins in each of the
original seven bags as \(x\). Therefore, the total number of coins
before finding the additional bag of 53 coins is \(7x\). After finding
the bag of 53 coins, the total number of coins becomes \(7x+53\). We
need to redistribute these coins into eight bags such that each bag has
the same number of coins. This means that \(7x+53\) must be divisible
by 8. Let's denote the number of coins in each of the eight bags after
redistribution as \(y\). Therefore, we have
\[
    7x+53=8y,
    \qquad
    7x=8y-53.
\]

We also know that the total number of coins after redistribution is more
than 200, so
\[
    8y>200,
    \qquad
    y>25.
\]

Since \(y\) must be an integer, the smallest possible value for \(y\)
is 26. Let's substitute \(y=26\) into the equation \(7x=8y-53\) and
solve for \(x\):
\[
    7x=8(26)-53,
    \qquad
    x=\frac{155}{7}.
\]

\textbf{Since 155 is not divisible by 7, \(x\) is not an integer, so
\(y=26\) is not a valid solution. Let's try the next smallest value for
\(y\), which is 27:}
\[
    7x=8(27)-53,
    \qquad
    x=\frac{163}{7}.
\]

\noindent(skip \ldots)

\textbf{Since 195 is not divisible by 7, \(x\) is not an integer, so
\(y=31\) is not a valid solution. Let's try the next smallest value for
\(y\), which is 32:}
\[
    7x=8(32)-53,
    \qquad
    x=\frac{203}{7}.
\]

\textbf{Since 203 is divisible by 7, \(x=29\) is an integer, so
\(y=32\) is a valid solution.} Therefore, the smallest number of coins
you could have had before finding the bag of 53 coins is
\(7x=7(29)=203\).

\textbf{Let's confirm this with Python code to ensure accuracy}
(skip \ldots)

\noindent\textbf{Output:} \texttt{203}

\noindent The smallest number of coins you could have had before finding
the bag of 53 coins is \(\boxed{203}\).

\end{reasonbox}

\captionof{figure}{MATH problem example: backtracking and
self-verification.}
\label{fig:math_example_5}

% ============================================================
% References begin on a new page
% ============================================================

\clearpage

% Start references on a new page after all examples.

%%%%%%%%%%%%%%%%%%%%%%%%%%%%%%%%%%%%%%%%%%%%%%%%%%%%%%%%%%%%

\end{document}